\documentclass[10pt,twocolumn,letterpaper]{article}

\usepackage{cvpr}              
\usepackage{bbding}
\usepackage{multirow} 
\usepackage{booktabs}
\usepackage{colortbl}
\usepackage{makecell}
\usepackage{diagbox}
\usepackage{xcolor}

\definecolor{cvprblue}{rgb}{0.21,0.49,0.74}
\usepackage[pagebackref,breaklinks,colorlinks,citecolor=cvprblue]
{hyperref}

\def\paperID{2358} 
\def\confName{CVPR}
\def\confYear{2024}

\title{Towards Anytime Retrieval: A Benchmark for Anytime Person Re-Identification}

\author{Xulin Li$^{1,2}$, Yan Lu$^{3}$, Bin Liu$^{1,2}$\thanks{Corresponding author.}, Jiaze Li$^{1,2}$, Qinhong Yang$^{1,2}$,\\ Tao Gong$^{1,2}$, Qi Chu$^{1,2}$, Mang Ye$^4$, Nenghai Yu$^{1,2}$\\ 
$^1$School of Cyber Science and Technology, University of Science and Technology of China\\ 
$^2$Anhui Province Key Laboratory of Digital Security\\ 
$^3$The Chinese University of Hong Kong\\ 
$^4$School of Computer Science, Wuhan University, China\\ lxlkw@mail.ustc.edu.cn, yanlu@cuhk.edu.hk, flowice@ustc.edu.cn, jz\_li@mail.ustc.edu.cn,\\ qhyang233@mail.ustc.edu.cn, \{tgong,qchu\}@ustc.edu.cn, yemang@whu.edu.cn, ynh@ustc.edu.cn}

\begin{document}
\maketitle

\begin{abstract}

In real applications, person re-identification (ReID) is expected to retrieve the target person at any time, including both daytime and nighttime, ranging from short-term to long-term.
However, existing ReID tasks and datasets cannot meet this requirement, as they are constrained by available time and only provide training and evaluation for specific scenarios.
Therefore, we investigate a new task called Anytime Person Re-identification (AT-ReID), which aims to achieve effective retrieval in multiple scenarios based on variations in time.
To address the AT-ReID problem, we collect the first large-scale dataset, AT-USTC, which contains 403k images of individuals wearing multiple clothes captured by RGB and IR cameras. 
Our data collection spans 21 months, and 270 volunteers were photographed on average 29.1 times across different dates or scenes, 4-15 times more than current datasets, providing conditions for follow-up investigations in AT-ReID.
Further, to tackle the new challenge of multi-scenario retrieval, we propose a unified model named Uni-AT, which comprises a multi-scenario ReID (MS-ReID) framework for scenario-specific features learning, a Mixture-of-Attribute-Experts (MoAE) module to alleviate inter-scenario interference, and a Hierarchical Dynamic Weighting (HDW) strategy to ensure balanced training across all scenarios.
Extensive experiments show that our model leads to satisfactory results and exhibits excellent generalization to all scenarios.
Our dataset and code are available at \href{https://github.com/kw66/AT-ReID}{https://github.com/kw66/AT-ReID.}

\end{abstract}
 
\begin{figure}[!t]
    \centering
    \includegraphics[width=1.0\linewidth]{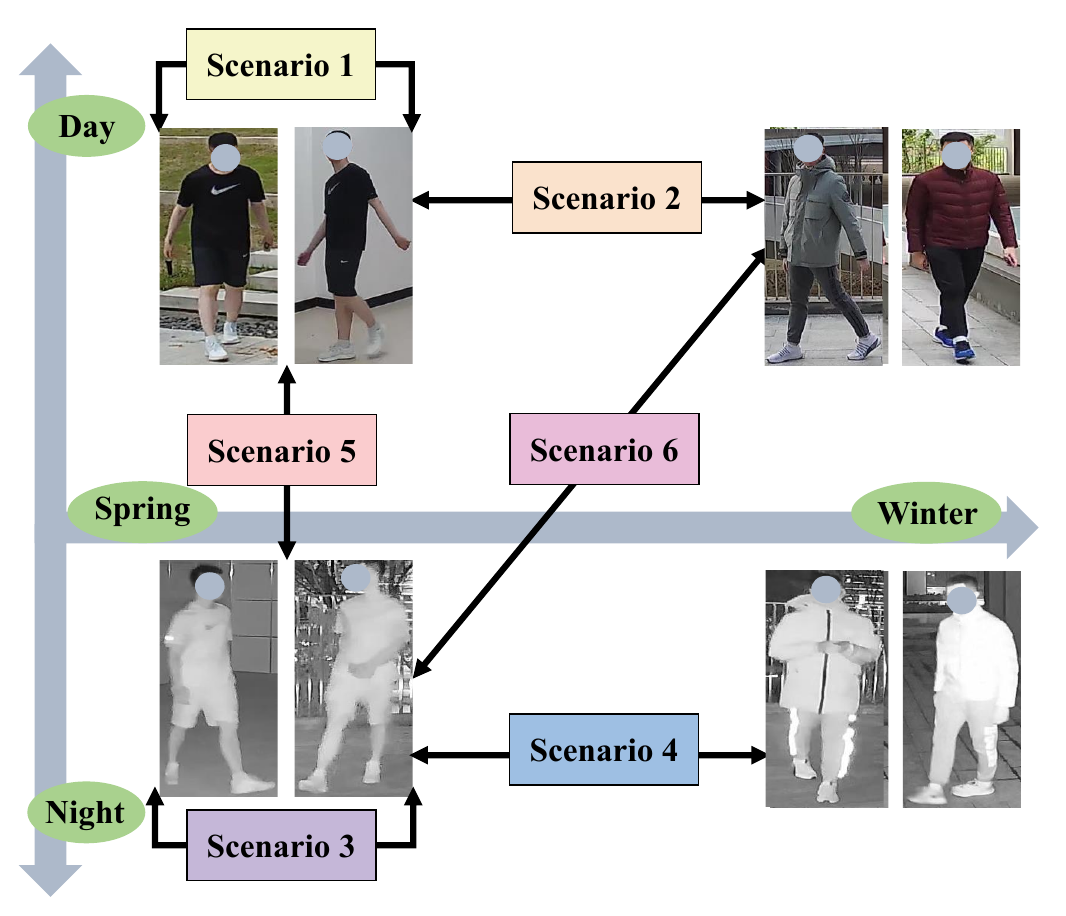}
    \vspace{-0.3cm}
    \caption{
    (a) AT-ReID aims to perform retrieval at any time, including both daytime and nighttime, ranging from short-term to long-term.
    }
    \vspace{-0.3cm}
    \label{fig:intro1}
\end{figure}

\section{Introduction}
\label{sec:intro}
Person re-identification (ReID) aims to retrieve specific pedestrians with given query images.
As illustrated in Fig.~\ref{fig:intro1}, a robust ReID system is expected to retrieve a person at any time, including daytime and nighttime, ranging from short-term to long-term, thereby satisfying the requirements of different surveillance scenarios.
This puts more challenges on the ReID system because the capturing time of the query image and the target image makes the task more variable.
For instance, if two images are captured during daytime and nighttime, respectively, they will have different modalities, and when there is a long time interval between their capturing, the person's appearance may change due to alterations in clothing.
Consequently, traditional ReID (Tr-ReID)~\cite{zheng2015scalable} may not perform effectively.

\begin{figure}[!t]
    \centering
    \includegraphics[width=1.0\linewidth]{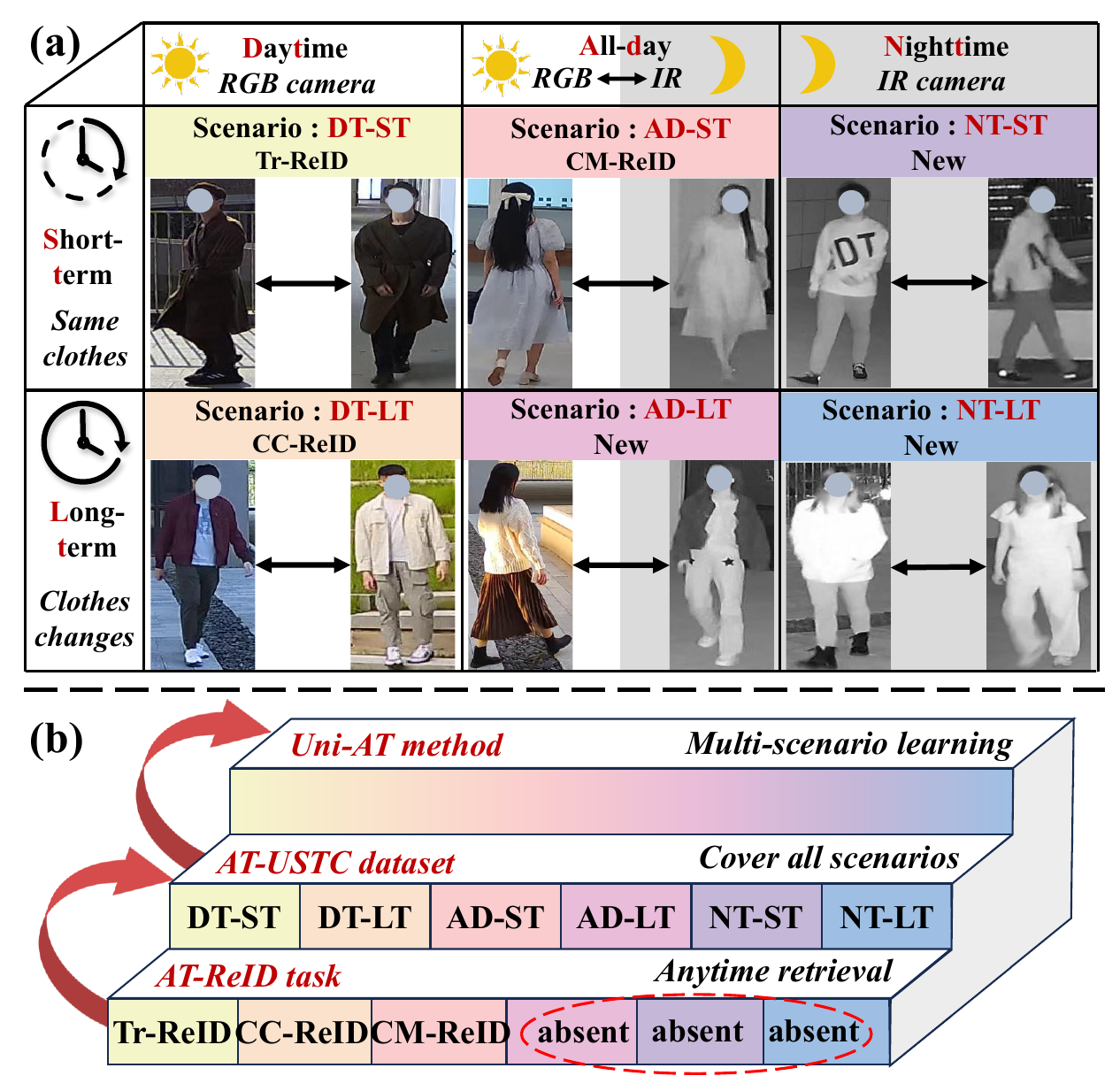}
    \vspace{-0.3cm}
    \caption{
    (a) Six non-overlapping scenarios based on variations in time. AT-ReID aims to perform retrieval in all of these scenarios.
    (b) Our solution of AT-ReID from the dataset to the model level.
    }
    \vspace{-0.3cm}
    \label{fig:intro2}
\end{figure}

The researchers acknowledged this challenge and attempted to address these problems separately.
They introduced the Visible-Infrared Cross-Modality ReID (CM-ReID)~\cite{wu2017rgb} to address the issue of searching between daytime RGB images and nighttime infrared (IR) images, and the Long-Term Cloth-Changing ReID (CC-ReID)~\cite{yang2019person} was proposed to handle long-term retrieval in which pedestrians change their clothes.
However, existing methods designed for these specific tasks were only able to achieve success in one of them and incapable of retrieving targets at any time simultaneously. 
This situation primarily arises from the absence of a long-term visible-infrared dataset covering all scenarios in Fig.~\ref{fig:intro2} (a), which should encompass diverse variations in clothing and modality for each individual.
The deficiency in intra-identity diversity of modalities and clothing in current ReID datasets has led to research gaps, especially in Nighttime Long-term (NT-LT) and All-day Long-term (AD-LT) scenarios.
Another issue arises from the poor generalization of task-specific methods in non-target scenarios. 
This is attributed to the differing learning objectives across different scenarios.
For instance, prior research~\cite{lu2020cross,qian2020long} has indicated that RGB-specific cues and clothing information are harmful to the All-day Short-term (AD-ST, CM-ReID) and Daytime Long-term (DT-LT, CC-ReID) scenarios, respectively, while they are crucial for the Daytime Short-term (DT-ST, Tr-ReID) scenario.

To meet the requirements of retrieving persons at any time, we investigate a new task called {\bf Anytime Person Re-identification (AT-ReID)} and propose to focus on its exploration from dataset to model level, as depicted in Fig.~\ref{fig:intro2} (b).
We collect the first corresponding large-scale dataset named {\bf AT-USTC}, which contains 403k images of 270 volunteers and covers all six scenarios in AT-ReID.
Our data collection spans 21 months, covering both day and night periods across the seasons of spring, summer, and winter.
We focus on simultaneously providing a greater variety of clothing and more RGB and IR cameras for each person.
Through efforts to expand in terms of capture dates, time periods, and scene variations, our AT-USTC provides a broader intra-identity diversity and more comprehensive AT-ReID cases than previous datasets.

To tackle the new challenge of multi-scenario retrieval in AT-ReID, we further propose a unified model named {\bf Uni-AT} to effectively handle all scenarios.
Given that the AT-ReID encompasses six different scenarios, the information shared among all scenarios becomes limited, and learning a unified representation for all scenarios is sub-optimal.
Therefore, we propose a novel Multi-Scenario ReID ({\bf MS-ReID}) framework with multiple classification tokens and a scenario-aware identity loss to facilitate effective learning of specific features for each scenario.
To achieve better discriminative feature extraction for different scenarios, we improve MS-ReID at both the model structure and optimization levels. 
Specifically, we propose a Mixture-of-Attribute-Experts ({\bf MoAE}) module, which builds the expert network and assigns different experts to address distinct scenarios, thus enabling the model to alleviate interference between scenarios.
Additionally, we define the attribute layer as the basic cell shared among experts with similar scenario attributes, e.g., DT-related attribute layers are shared among DT-LT and DT-ST experts.
With this, the model can benefit from multiple interrelated scenarios.
And we propose a Hierarchical Dynamic Weighting ({\bf HDW}) strategy, that tackles the AT-ReID training from the multi-task learning view.
It establishes all scenarios into several tasks and balances the training for different tasks with a loss weighting scheme.
This method considers multiple relevant tasks when computing weights, implicitly modeling the relationships between tasks and leading to better optimization of the overall multi-scenario learning framework.

Our main contributions can be summarized as follows:

\noindent $\bullet$ 
We investigate a new task called AT-ReID, which aims at enabling retrieval at any time moment and across different time intervals.
We contribute for the first time a large-scale dataset named AT-USTC to support the study of AT-ReID.
Compared to existing datasets, AT-USTC stands out for its long data collection period and the inclusion of both RGB and IR camera footages, meeting the requirement of AT-ReID.
Importantly, our data collection has obtained the consent of each volunteer.

\noindent $\bullet$ 
We propose a Uni-AT model to effectively handle all scenarios of AT-ReID.
In Uni-AT, three components, a new multi-scenario ReID framework, a Mixture-of-Attribute-Experts module, and a Hierarchical Dynamic Weighting training strategy are proposed to tackle the new challenges of multi-scenario retrieval in AT-ReID tasks.
Extensive experiments show that our model leads to satisfactory results and exhibits excellent generalization to all scenarios.

\section{Related Work} 
\label{sec:relate}
\paragraph{Person Re-Identification.}
Traditional ReID (Tr-ReID) aims to achieve short-term pedestrian retrieval in the RGB modality.
The corresponding datasets, such as Market1501~\cite{zheng2015scalable}, CUHK03~\cite{li2014deepreid}, and MSMT17~\cite{wei2018person}, focused on providing more identities as well as more camera variations.
Tr-ReID methods involve general pedestrian retrieval techniques, such as the design of more robust backbone networks~\cite{luo2019bag, ye2021deep, he2021transreid}, effective ReID loss functions~\cite{sun2020circle}, and the utilization of part-level features~\cite{sun2018beyond} to achieve discriminative representations of pedestrians. 

Visible-Infrared Cross-Modality ReID (CM-ReID) aims to achieve short-term pedestrian retrieval between the RGB and the infrared (IR) modalities.
The corresponding datasets, such as SYSU-MM01~\cite{wu2017rgb}, RegDB~\cite{nguyen2017person}, and LLCM~\cite{zhang2023diverse}, focused on providing more RGB and IR cameras.
Some CM-ReID methods~\cite{feng2023shape,yang2023towards} aimed to project features from different modalities into the same feature space, while others~\cite{ye2020dynamic,lu2020cross,li2022counterfactual,fang2023visible} aimed to learn cross-modality relationships.

Long-term Cloth-changing ReID (CC-ReID) aims to achieve long-term pedestrian retrieval in the RGB modality.
The corresponding datasets, such as PRCC~\cite{yang2019person}, LTCC~\cite{qian2020long}, and DeepChange~\cite{xu2023deepchange}, focused on providing clothing variations for each person.
Some CC-ReID methods~\cite{chen2021learning,jin2021cloth,liu2023dual,guo2023semantic} introduced additional data such as contour, key points, human parsing, and 3D shape for model training, while others~\cite{gu2022clothes,han2023clothing,yang2023good,li2023clothes} utilized RGB images only to learn robust clothing-irrelevant feature.

The aforementioned tasks and datasets can only cover a portion of the AT-ReID scenarios.
In addition, some unified methods~\cite{chen2023towards,tang2023humanbench,ci2023unihcp,he2023retrieve,li2024all,zheng2024versatile} focus on multiple ReID tasks, such as text/sketch-to-RGB ReID, clothes template based CC-ReID, and occlusion ReID, as well as human-centric tasks, such as human parsing, pose estimation, and pedestrian detection.
Our research is distinct from previous methods as it is the first to focus on the availability of ReID at any time and proposes a relevant dataset and method to bridge the gap between existing research and AT-ReID.

\paragraph{Multi-Task Learning.}
Multi-task learning (MTL) refers to building a model that can handle multiple distinct tasks~\cite{ben2008notion,meyerson2017beyond}.
By sharing parameters between tasks, MTL methods achieve efficient memory and data utilization and expect to derive benefits from multiple related tasks.
In AT-ReID, various input modalities and learning objectives are present in different scenarios. 
Retrieval in each scenario can be considered an individual ReID task, and it is promising that employing MTL methods can improve the overall efficacy of the model across all scenarios.

Some MTL methods focused on network architecture~\cite{ma2019snr,liu2019end,zhu2022uni,wang2023multitask} to achieve more effective parameter sharing.
Recently, some effective approaches~\cite{ma2018modeling,tang2020progressive,zhu2022uni,bao2022vlmo,rajbhandari2022deepspeed} are to utilize the Mixture-of-Experts (MoE)~\cite{jacobs1991adaptive} model that employs multiple expert sub-networks to tackle multi-task learning.
Compared to these MoE methods, our MoAE constructs scenario experts in a more flexible sharing manner, making the model benefit from multiple interrelated scenarios.
Other methods focused on MTL optimization, such as manipulating gradient~\cite{yu2020gradient,liu2021conflict,liu2021towards} and adjusting the loss weight by task difficulty, training speed, and priority~\cite{guo2018dynamic,kendall2018multi,liu2019end,chen2023towards}.
Our HDW method groups tasks based on their attributes and applies hierarchical dynamic weighting to the loss of each task, achieving a more effective task balance.

\section{AT-USTC Dataset}
\label{sec:dataset}

\begin{figure}[t]
    \centering
    \includegraphics[width=1.0\linewidth]{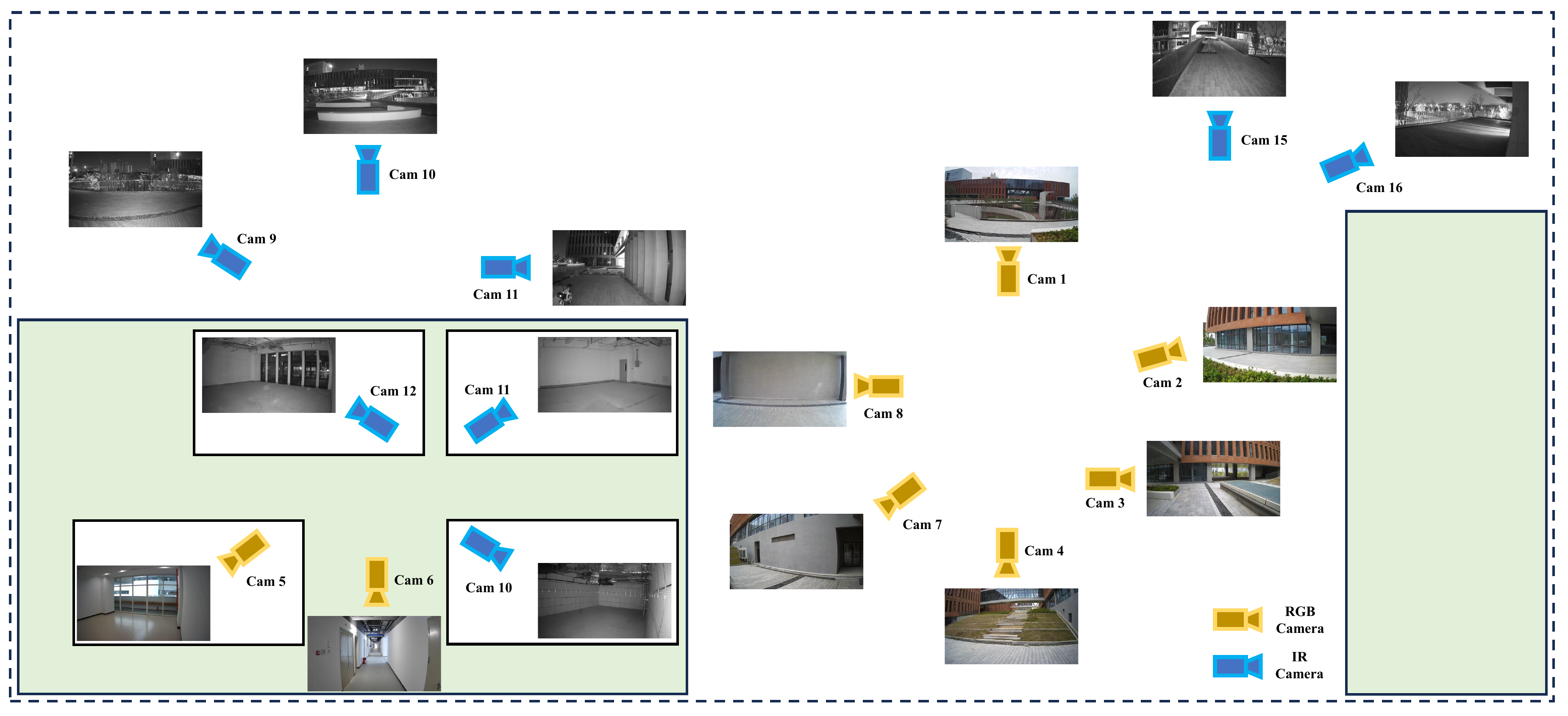}
    \vspace{-0.3cm}
    \caption{
    The plan of the camera layout for collecting data.
    }
    \vspace{-0.3cm}
    \label{fig:place}
\end{figure}

\begin{figure*}[!t]
    \centering
    \includegraphics[width=1.0\linewidth]{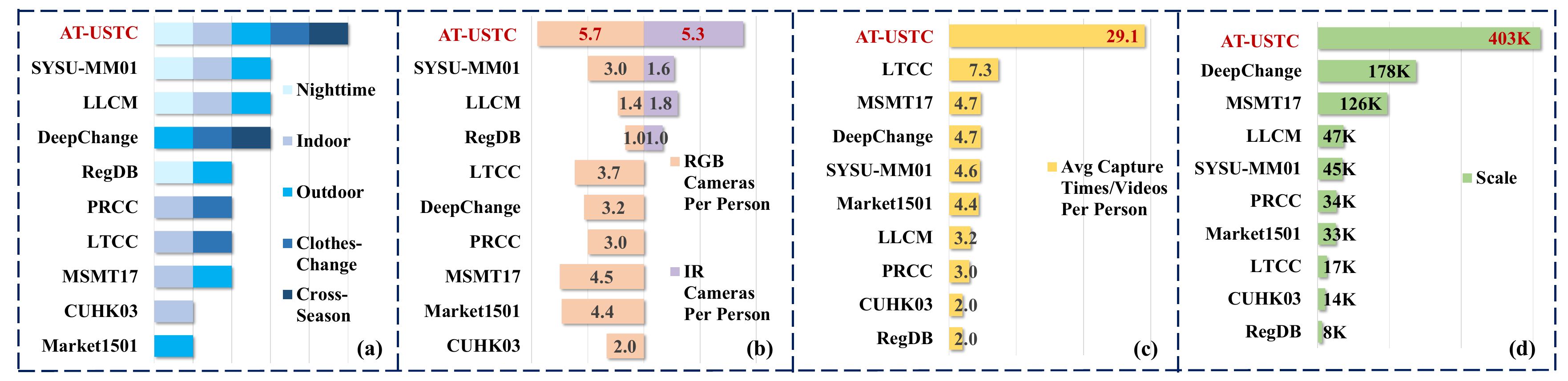}
    \vspace{-0.4cm}
    \caption{
    Statistics of our AT-USTC and several popular ReID datasets.}
    \vspace{-0.3cm}
    \label{fig:stat}
\end{figure*}

\begin{figure}[t]
    \centering
    \includegraphics[width=1.0\linewidth]{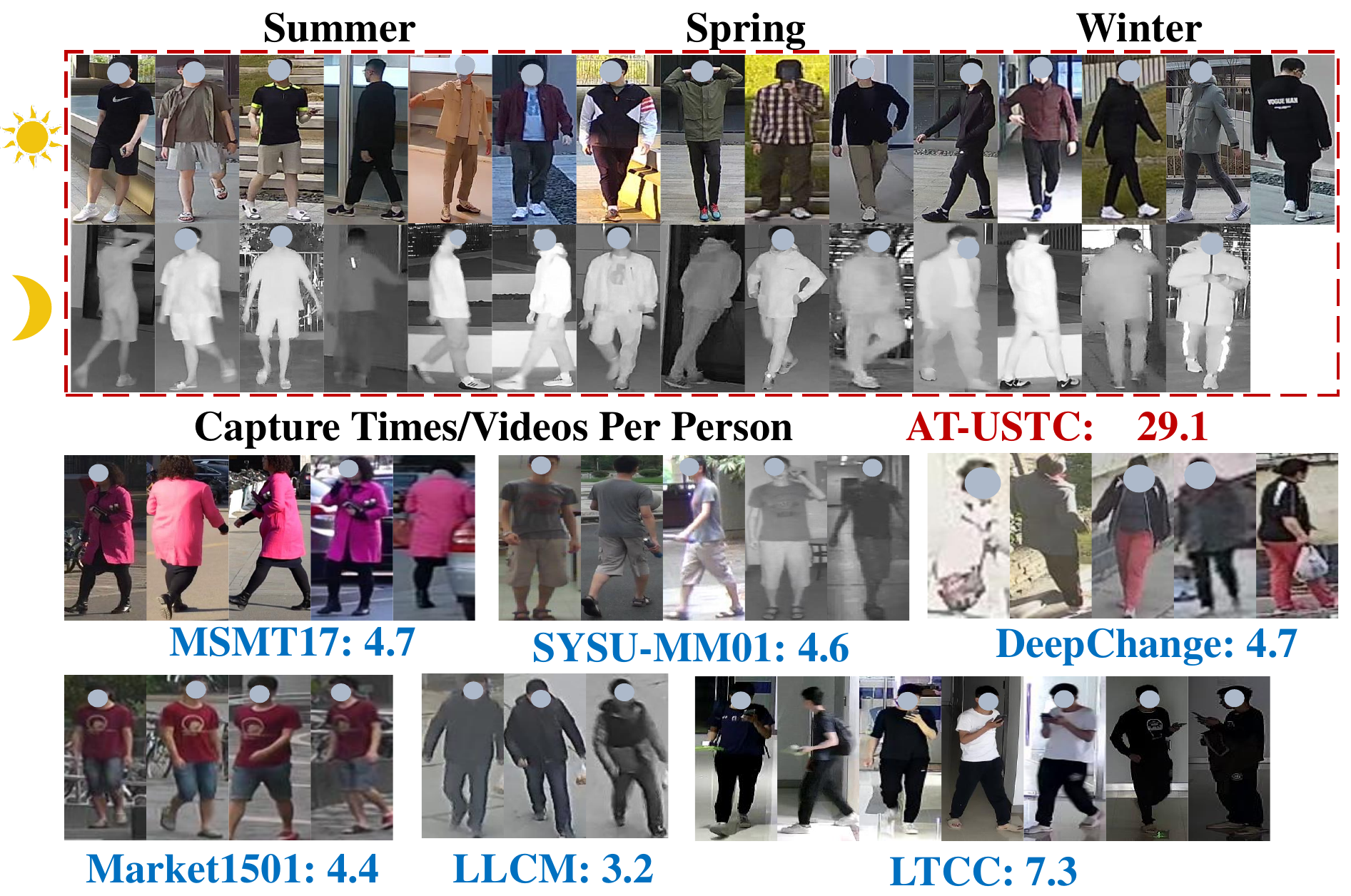}
    \vspace{-0.3cm}
    \caption{
    Each person in AT-USTC has been photographed 29.1 times on average, resulting in higher intra-identity diversity. Images of each dataset all belong to the same individual.
    }
    \vspace{-0.3cm}
    \label{fig:diver}
\end{figure}

\paragraph{Dataset Description.}
AT-USTC is the first AT-ReID benchmark that includes 403,599 (199,803 RGB and 203,796 IR) images of 270 identities and 710 sets of different clothing captured by 16 cameras.
As shown in Fig.~\ref{fig:place}, we deployed 8 RGB and 8 IR cameras across 16 non-overlapping locations, comprising 5 indoor and 11 outdoor scenes.
We filmed videos spans 21 months including spring, summer, and winter, with temperatures ranging from -3$^{\circ}$C to 33$^{\circ}$C to cover a wider range of clothing types.
Each individual in our training set has 2-14 outfits with an average of 3.6, which facilitates retrieval in long-term scenarios.
Due to the variations in both modality and clothing in AT-USTC, the process of capturing and annotating the data is more time-consuming compared to other datasets.
We made considerable effort to provide annotations, including labels for person, camera, and clothing.

\paragraph{Privacy Protection.}
Following the established ReID datasets~\cite{wu2017rgb,yang2019person}, we made efforts for privacy protection in five aspects: 
1) Data collection was authorized by the relevant authorities, involving the deployment of cameras and image capture.
2) The individuals we photographed did not include minors.
3) Each volunteer who participated in the filming has signed a standardized consent agreement, agreeing to the release of their images for academic research purposes.
4) Our AT-USTC dataset does not include any personal information beyond the captured images.
5) Individuals or organizations seeking to use our dataset are required to sign the corresponding dataset release agreement, which imposes restrictions on the dataset's copyright, usage, modification, and redistribution.

\vspace{-2mm}
\paragraph{Dataset Advantages.}
As shown in Fig.~\ref{fig:stat} (a), our AT-USTC captured images in various scenarios, including daytime, nighttime, indoor, outdoor, and cross-season intervals, providing comprehensive variations of modality, clothing, camera, and scene.
In contrast, the three Tr-ReID datasets, Market1501~\cite{zheng2015scalable}, CUHK03~\cite{li2014deepreid}, and MSMT17~\cite{wei2018person} do not include clothing changes and IR cameras; the three CM-ReID datasets, SYSU-MM01~\cite{wu2017rgb}, RegDB~\cite{nguyen2017person}, and LLCM~\cite{zhang2023diverse} do not consider clothing changes; the three CC-ReID datasets, PRCC~\cite{yang2019person}, LTCC~\cite{qian2020long}, and DeepChange~\cite{xu2023deepchange} do not include IR cameras.
Additionally, certain datasets, such as synthetic datasets~\cite{wan2020person}, datasets from movies \protect\cite{shu2021large}, and the internet \protect\cite{yildiz2024entire}, differ significantly from the surveillance environment domain and are therefore not within the scope of comparison.

As shown in Fig.~\ref{fig:stat} (b), we provide rich camera variations for each person during day and night, facilitating cross-camera and cross-modality retrieval.
The number of cameras per identity~\cite{song2025exploring} reflects the camera diversity of the dataset.
On average, each person of AT-USTC appears in 5.7 RGB and 5.3 IR cameras, totaling 11 cameras, which is significantly higher than other datasets.

As illustrated in Fig.~\ref{fig:stat} (c) and Fig.~\ref{fig:diver}, each person in AT-USTC has been photographed 29.1 times on average, and the identities in the training set were photographed an average of 40.0 times.
Therefore, this results in significantly higher intra-identity diversity and visual variations of our AT-USTC dataset. Compared with existing datasets, our AT-USTC exhibits day and night photography, diverse cameras, and captures across multiple seasons, enabling it to encompass all scenarios of AT-ReID.

As shown in Fig.~\ref{fig:stat} (d), the scale of our AT-USTC significantly exceeds other datasets. This is due to the higher intra-identity diversity exhibited by each individual within the dataset. In addition to the rich variations in cameras, modalities, and clothing, the average duration of each video in our dataset is 50 seconds, resulting in greater diversity of postures.

\vspace{-2mm}
\paragraph{Data Split.}
We have a fixed split of the dataset into training and testing sets.
The training set consists of 135 people with 286,087 images, and the testing set consists of another 135 people with 117,512 images.
We partitioned 20\% (55,060) images from the training set for validation purposes.
Existing datasets primarily evaluate a single scenario, while we construct separate gallery sets and query sets for all six scenarios covered by AT-ReID to facilitate a comprehensive assessment of model performance and explore anytime retrieval.
For each identity, we selected three query images and three gallery images from video clips featuring the same identity, captured by the same camera, and with the same clothing. Under this configuration, the gallery contains an average of approximately 25 images per identity.
 
\begin{figure*}[!t]
    \centering
    \includegraphics[width=1.0\linewidth]{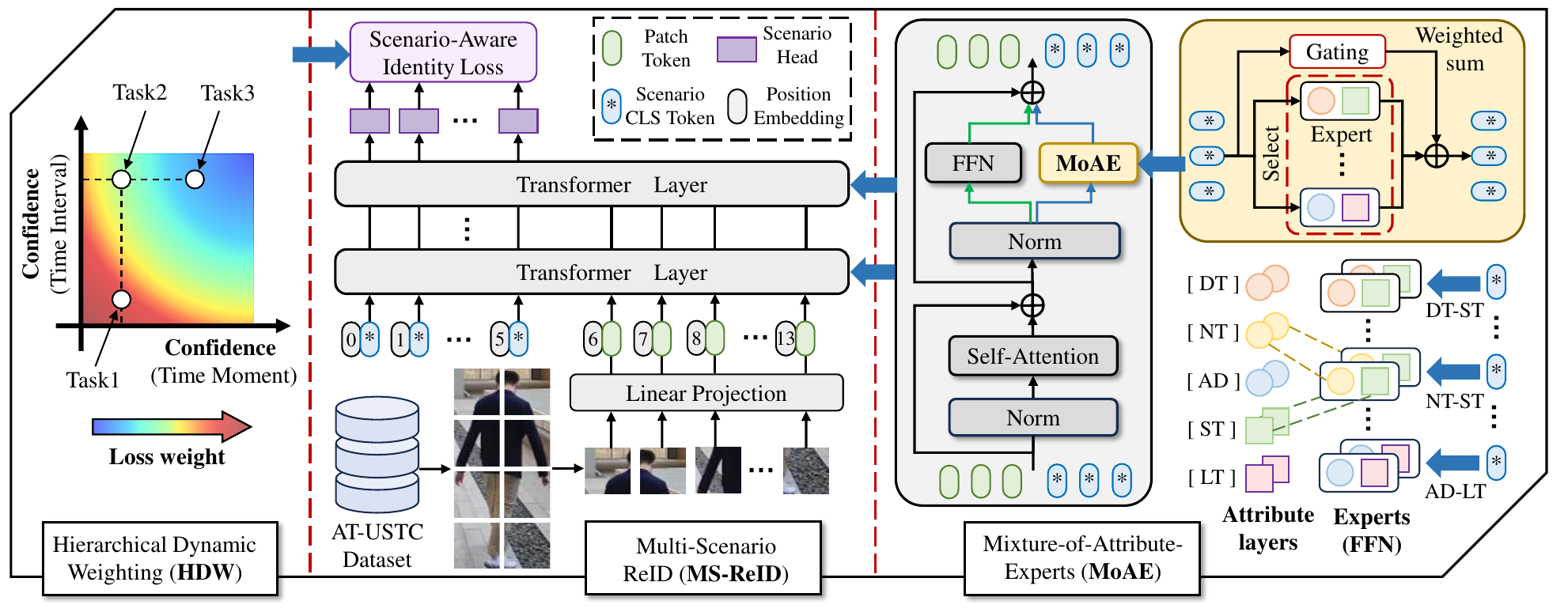}
    \vspace{-0.2cm}
    \caption{
    The pipeline of our Uni-AT.
    The DT, NT, AD, ST, and LT denote daytime, nighttime, all-day, short-term, and long-term cases, respectively.
    With a Multi-Scenario ReID framework, a Mixture-of-Attribute-Experts module, and a Hierarchical Dynamic Weighting scheme, Uni-AT enhances the learning of diverse scenario-specific features and improves model generalization.
    }
    \vspace{-0.3cm}
    \label{fig:method}
\end{figure*}

\section{Method}  
\label{sec:method}

\paragraph{Overview.} Anytime ReID (AT-ReID) aims to perform retrieval in multiple scenarios, including daytime short-term (DT-ST), daytime long-term (DT-LT), all-day short-term (AD-ST), all-day long-term (AD-LT), nighttime short-term (NT-ST), and nighttime long-term (NT-LT) scenarios.

The pipeline of our proposed Unified AT-ReID model (Uni-AT) is shown in Fig.~\ref{fig:method}. 
The image is fed into a Multi-Scenario ReID (MS-ReID) framework to extract several types of scenario features for accurate retrieval in all covered scenarios of AT-ReID.
To treat each scenario optimally, we further propose a Mixture-of-Attribute-Experts (MoAE) module to effectively capture scenario-specific clues and mitigate inter-scenario interference.
To balance feature learning in different scenarios, we proposed a Hierarchical Dynamic Weighting (HDW) scheme to train the whole model more effectively in an end-to-end way. 

\subsection{Multi-Scenario ReID}
\label{sec:msreid}
\paragraph{Model Architecture.} 
The proposed Multi-Scenario ReID (MS-ReID) framework is designed to extract multiple scenario-specific features more effectively.
We choose Vision Transformer (ViT)~\cite{dosovitskiy2020image} as our backbone.
The input image $x_i$ is split into patches and mapped to patch tokens.
Then we establish 6 CLS tokens $t_i^s$ to extract image features and assign each one with a corresponding scenario $s$.
These CLS tokens serve as information gatherers, collecting different scenario-specific knowledge from patch tokens by stacked self-attention modules.

Note that the main principle of our MS-ReID is to extract different features for different scenarios separately rather than use a single unified representation for all scenarios because the latter sacrifices specific clues in each scenario.
The main concern is based on a prior that scenario-specific information can lead to optimal results under specific cases.
For example, color information is suitable for daytime retrieval and clothes cues are discriminative for short-term situations. 
Moreover, AT-ReID is a scenario-determinable task, as the practical ReID involves retrieving between the query image and the gallery images in the video surveillance, and the shooting timestamps can be easily accessed.
When faced with an uncertain scenario, the default is set to all-day/long-term to account for potential modality variations/clothing changes.
Thus, our framework is adaptable and capable of offering more precise solutions for determined scenarios.

\paragraph{Scenario-aware identity loss.} 
The common identity loss provides undifferentiated supervision for all scenarios, which can only learn the shared information across all scenarios but cannot capture scenario-specific cues.
Therefore, we propose a scenario-aware identity loss $\mathcal{L}^s_{id}$ for our MS-ReID framework to supervise feature learning for each covered scenario, where different scenarios have non-shared classifiers, distinct negative category sets, and different modality filtering mechanisms. $\mathcal{L}^s_{id}$ is derived as follow:

\begin{equation}
\label{eq:ce1}
\begin{aligned}
    \mathcal{L}^s_{id}(t^s_i) = 
    -\log(p^{s}_{i,gt}), 
\end{aligned}
\end{equation}
where $p^{s}_{i,gt}$ is the classification probability for scenario $s$. $p^{s}_{i,gt}$ is derived as follow:
\begin{equation}
\label{eq:ce2}
\begin{aligned}
    p^{s}_{i,gt} = 
    \frac{\exp(o^{s}_{i,gt})}{\exp(o^{s}_{i,gt})+\sum_{j\in N_{s}} \exp(o^{s}_{i,j})},
\end{aligned}
\end{equation}
where $o_i^{s}$ is classification logit generated from the scenario CLS token $t_i^s$ and $N_{s}$ is the negative category set for the scenarios $s$. 
We employ distinct $N_{s}$ for ST-related and LT-related scenarios.
Specifically, for ST cases, we treat different clothes as different categories in classification to guide the model to attend to fine-grained discriminative information about clothes.
However, we do not expect the model to classify the same person's images with different clothes into different categories because they share consistent semantic body information. 
To achieve this goal, we set $N_{s}$ of ST cases as the clothes ID set while each clothes in this set has different owners with the ground-truth clothes.
For the LT cases, we follow the traditional ReID setting to define the category space as the set of person IDs.
So $N_{s}$ of LT cases are different ID persons directly. 
In addition to distinguishing between ST and LT scenarios, scenario-aware identity loss also includes a modality filtering mechanism.
For RGB images, we supervise the DT-ST, DT-LT, AD-ST, and AD-LT tokens of their potential scenarios while neglecting the NT-related ones that are not available to RGB images. 
Similarly, for IR images, we ignore their DT-related tokens.

Our MS-ReID can fit the goal of each scenario separately, facilitating multi-scenario learning on a dataset with modality and clothing variations.

\subsection{Mixture-of-Attribute-Experts}
\label{sec:moae}
In our MS ReID framework, CLS tokens are supervised by distinct identity loss, aiming to extract corresponding scenario-specific features.
However, similar to the case of multi-task learning~\cite{meyerson2017beyond}, the parameter-shared ViT feature extraction network may result in potential gradient conflicts.
For instance, the all-day scenario focuses solely on modality-shared information while the daytime scenario needs to also consider RGB-specific information, leading to mutual interference in feature learning between scenarios.

To tackle this problem, inspired by Mixture-of-Experts (MoE) methods~\cite{jacobs1991adaptive,ma2018modeling}, we propose a novel Mixture-of-Attribute-Experts (MoAE) module.
As depicted in Fig.~\ref{fig:method}, the MoAE module is added parallel with the feed-forward network (FFN) in the transformer layers.
In each transformer layer, patch tokens are fed into the original FFN, while the CLS tokens are fed into the MoAE module.
We establish MoAE with $6n$ experts and assign different experts to address distinct scenarios. 
For each scenario $s$, there are $n$ specific experts $\{E^s_j\}_{j=1}^{n}$, where $E^s_j$ represents a single-layer FFN comprising two linear layers and a gelu~\cite{hendrycks2016gaussian} activation function.
Given an input CLS token $t_i^s \in R^d$ corresponding to the scenario $s$, the MoAE selects and combines experts through a gating network, producing the output $y \in R^d$. More precisely, $y$ is the weighted sum of the outputs from the $n$ experts:
\begin{equation}
\label{eq:moe}
\begin{aligned}
    y = \sum\nolimits_{j=1}^n G^s(t_i^s)_j E_j^s(t_i^s),
\end{aligned}
\end{equation}
where $G^s(t_i^s) \in R^n$ is the weight of $n$ experts, calculated by a gating network:
\begin{equation}
\label{eq:gate}
\begin{aligned}
    G^s(t_i^s) = top_k(\text{softmax}(W_g^s\cdot t_i^s)),
\end{aligned}
\end{equation}
where, $W_g^s \in R^{n\times d}$ is the trainable weights for scenario $s$ in gate decision, and the $top_k(\cdot)$ operator sets all values to zeros except the top-$k$ largest values. 
Except for the selected $k$ experts, others do not need to be computed for saving computation.
Following~\cite{ma2018modeling,tang2020progressive,bao2022vlmo}, we set $k$ to 1 to obtain sparse and efficient expert networks.
With this module, different inputs can utilize experts in distinct ways and capture scenario-specific clues.

Additionally, we note the existence of potential relationships among various scenarios.
For example, DT-ST and DT-LT scenarios both have the ``DT" attribute, indicating that they require the model to extract RGB-specific features.
To introduce this knowledge and establish relationships across scenarios, we construct five types of attribute layers as the basic cells shared among experts with similar scenario attributes. 
Among these, ``DT", ``AD", and ``NT" attribute layers belong to the category of Time-Moment (TM), while ``ST", and ``LT" attribute layers belong to the category of Time-Interval (TI). 
In practice, attribute layers are linear layers and we derive the experts by combining one TM attribute layer and one TI attribute layer as follows:
\begin{equation}
\label{eq:exp}
\begin{aligned}
    E^s(\cdot) = a_1(gelu(a_2(\cdot))),
\end{aligned}
\end{equation}
where $a_1$ and $a_2$ are the attribute layers associated with scenario $s$.
In summary, our attribute layers are shared across different scenarios, which can serve as prior knowledge for the relationships between scenarios. 
Compared to other MoE methods~\cite{ma2018modeling,tang2020progressive,bao2022vlmo}, our MoAE is more flexible and effective for expert construction, which lets the model benefit from multiple interrelated scenarios.

\begin{table*}[t]
    \renewcommand{\arraystretch}{1.4} 
    \fontsize{9}{9}\selectfont
    \setlength\tabcolsep{5.0pt}
    \centering
    \begin{tabular}{l c c c c c c c c c c c c c}
        \toprule[1pt]&
        \multicolumn{2}{c}{\cellcolor[HTML]{F5F5CA}{Market1501}}&
        \multicolumn{2}{c}{\cellcolor[HTML]{F5F5CA}{CUHK03}}&
        \multicolumn{2}{c}{\cellcolor[HTML]{FACCCE}{SYSU-MM01}}&
        \multicolumn{2}{c}{\cellcolor[HTML]{FAE2CC}{PRCC}}&
        \multicolumn{2}{c}{\cellcolor[HTML]{FAE2CC}{LTCC}}&
        \multicolumn{2}{c}{\cellcolor[HTML]{C9FAFC}{AVG}}
        \cr\cmidrule(r){2-3}\cmidrule(r){4-5}\cmidrule(r){6-7}
        \cmidrule(r){8-9}\cmidrule(r){10-11}\cmidrule(r){12-13}
        \multirow{-2}*{\diagbox[width=10em]{Train}{Test}}&Rank-1&mAP&Rank-1&mAP&Rank-1&mAP
        &Rank-1&mAP&Rank-1&mAP&Rank-1&mAP
        \cr\midrule
        \cellcolor[HTML]{F5F5CA}{MSMT17 (1041 IDs)}&57.63&30.08&14.64&13.81&
        4.49&6.81&24.41&23.33&20.66&8.53&
        24.37&16.51\cr
        \cellcolor[HTML]{FACCCE}{LLCM (713 IDs)}&46.79&19.79&4.36&4.89&
        7.17&8.78&33.45&26.17&12.76&5.93&
        20.91&13.10\cr
        \cellcolor[HTML]{FAE2CC}{DeepChange (450 IDs)}&57.48&30.42&6.00&6.11&
        3.60&5.85&25.52&24.12&15.31&5.90&
        21.58&14.48\cr\midrule
        \cellcolor[HTML]{C9FAFC}{AT-USTC (135 IDs)}&
        \textbf{60.30}&\textbf{34.95}&\textbf{21.71}&
        \textbf{21.35}&\textbf{26.49}&\textbf{25.61}&
        \textbf{42.87}&\textbf{36.81}&\textbf{25.00}&
        \textbf{9.97}&\textbf{35.27}&\textbf{25.74}\cr
        \bottomrule[1pt]
    \end{tabular}
    \vspace{-0.1cm}
    \caption{
    Cross-domain generalization experiments in different datasets. Rank-1 accuracy (\%) is reported.
    }
    \vspace{-0.1cm}
    \label{tab:dataset}
\end{table*}

\subsection{Hierarchical Dynamic Weighting}
\label{sec:hdw}

In our MS-ReID framework, we utilize scenario-aware identity loss to supervise each scenario, which can be considered as multi-task learning.
Through joint learning across multiple tasks, we aim to increase the model's generalization ability in various scenarios.
However, different tasks have different levels of difficulty and learning curves.
Simply summing all losses with fixed weights to optimize the overall framework may not provide adequate training for some tasks, leading to limited robustness.

To achieve a balanced optimization across all tasks, we propose a Hierarchical Dynamic Weighting (HDW) scheme.
The idea of HDW is that, when some tasks retrieve corresponding images with low predicted confidence, they should contribute more to the final loss, and vice versa.
The HDW can be exported as follows:
\begin{equation} 
\label{eq:loss} 
\begin{aligned}
    \mathcal{L}_{total} = \sum\nolimits_{s\in S} w^s\cdot\mathcal{L}^{s}_{id},
\end{aligned} 
\end{equation}
where $w^s$ is the weight of the scenario $s$. 
Additionally, we observe that the learning of different tasks is interrelated rather than independent.
For instance, the retrieval task in DT-ST scenarios requires RGB-specific features and clothing features. The former can be learned from two DT-related tasks, while the latter can be learned from three ST-related tasks. 
Therefore, the weight adjustment for these tasks should also adhere to this principle. 
To achieve this, $w^s$ is calculated by two terms and derived as follows:
\begin{equation} 
\label{eq:weight0} 
\begin{aligned}
w^s = w^s_{tm}\cdot w^s_{ti}.
\end{aligned} 
\end{equation}
Inspired by the Focal Loss~\cite{lin2017focal}, $w^s_{tm}$ and $w^s_{ti}$ can be computed as follows:
\begin{equation} 
\label{eq:weight} 
\begin{aligned}
w^s_{tm} = (1-p^{s}_{tm})^\frac{1}{2}, w^s_{ti} = (1-p^{s}_{ti})^\frac{1}{2},
\end{aligned} 
\end{equation}
where $p^{s}_{tm}$ and $p^{s}_{ti}$ corresponding time moment and time interval confidences for scenario $s$, respectively.
For example, if $s$ is the DT-LT scenario, $p^{s}_{tm}$ is the confidence of the DT situation while $p^{s}_{ti}$ is the LT case confidence.
We compute the confidence by averaging the ground truth classification probabilities of CLS tokens from scenarios $s'$ in each training batch:
\begin{equation}  
\label{eq:confidence}  
\begin{aligned} 
p^{s}_{tm} &= mean(exp(-L_{id}^{s'}(t^{s'}_i))),\quad s_{tm}' = {s}_{tm},\\
p^{s}_{ti} &= mean(exp(-L_{id}^{s'}(t^{s'}_i))),\quad s_{ti}' = {s}_{ti},
\end{aligned}  
\end{equation}
where scenarios $s'$ carry the same time moment/time interval attribute as target scenario $s$.

Different from traditional multi-task learning strategies~\cite{guo2018dynamic} assuming each task is independent of the other, our joint weighting method considers multiple relevant tasks when computing weights, implicitly modeling the relationships between tasks and leading to better optimization of the overall multi-task learning framework.

\section{Experiments}
\label{sec:exper}

\begin{table}[t]
    \fontsize{9}{9}\selectfont     
    \setlength\tabcolsep{5.0pt}
    \renewcommand{\arraystretch}{1.4}
    \centering
    \begin{tabular}{l c c c c c}
        \toprule[1pt]
        Method&Experts&Params&Time&Rank-1&mAP
        \cr\midrule
        MS-ReID&0&1.00\texttimes&1.00\texttimes
        &52.03&37.49\cr\midrule
        \multirow{2}*{+ MMoE~\cite{ma2018modeling}}
        &6&4.90\texttimes&1.60\texttimes
        &53.00&38.22\cr			 
        &12&8.80\texttimes&2.18\texttimes
        &53.17&38.62\cr\midrule
        \multirow{3}*{+ PLE~\cite{tang2020progressive}}
        &6&4.90\texttimes&1.14\texttimes         
        &52.77&38.44\cr	 
        &7&5.56\texttimes&1.26\texttimes
        &53.17&38.54\cr	 
        &12&8.80\texttimes&1.81\texttimes
        &53.20&38.67\cr\midrule
        \multirow{3}*{+ VLMo~\cite{bao2022vlmo}}
        &3&2.95\texttimes&1.07\texttimes         
        &52.81&38.10\cr
        &6&4.90\texttimes&1.25\texttimes         
        &53.30&38.59\cr
        &12&8.80\texttimes&1.53\texttimes
        &53.40&38.57\cr\midrule
        \multirow{2}*{+ MoAE (ours)}
        &6&2.62\texttimes&1.06\texttimes
        &52.98&38.40\cr
        &12&4.25\texttimes&1.20\texttimes
        &\bf{53.70}&\bf{38.76}\cr
        \bottomrule[1pt]
    \end{tabular}
    \vspace{-0.1cm}
    \caption{Comparison with other MoE methods on AT-USTC.}
    \vspace{-0.1cm}
    \label{tab:moe}
\end{table}

\paragraph{Datasets.}
We primarily conducted experiments on the proposed AT-USTC dataset due to the lack of support for multi-scenario training and evaluation of AT-ReID in existing datasets.
To further evaluate our AT-USTC dataset and Uni-AT method, we utilized several popular ReID datasets, including MSMT17~\cite{wei2018person}, Market1501~\cite{zheng2015scalable}, CUHK03~\cite{li2014deepreid}, SYSU-MM01~\cite{wu2017rgb}, LLCM~\cite{zhang2023diverse}, DeepChange~\cite{xu2023deepchange}, PRCC~\cite{yang2019person}, and LTCC~\cite{qian2020long}, 
for cross-domain generalization experiments.

\paragraph{Evaluation Protocols.}
The Rank-$k$ matching accuracy and mean average precision (mAP) are adopted as evaluation metrics.
For AT-USTC, we conducted separate tests in six different scenarios, including DT-ST, DT-LT, NT-ST, NT-LT, AD-ST, and AD-LT.
The average performance of six scenarios is referred to as {\bf Any-Time}, which evaluates the model's ability to retrieve at any given time.
For other datasets, we adhered to the evaluation settings of their original papers.

\begin{table*}[t]
    \renewcommand{\arraystretch}{1.4}      
    \fontsize{9}{9}\selectfont     
    \setlength\tabcolsep{4.0pt}
    \centering
    \begin{tabular}{ccccccccccccccccc}
        \toprule[1pt]
        \multicolumn{3}{c}{Method}&
        \multicolumn{2}{c}{\cellcolor[HTML]{C9FAFC}{Any-Time}}&
        \multicolumn{2}{c}{\cellcolor[HTML]{F5F5CA}{DT-ST}}&
        \multicolumn{2}{c}{\cellcolor[HTML]{FAE2CC}{DT-LT}}&
        \multicolumn{2}{c}{\cellcolor[HTML]{C5B7D8}{NT-ST}}&
        \multicolumn{2}{c}{\cellcolor[HTML]{9EBFE3}{NT-LT}}&
        \multicolumn{2}{c}{\cellcolor[HTML]{FACCCE}{AD-ST}}&
        \multicolumn{2}{c}{\cellcolor[HTML]{E9BCD9}{AD-LT}}
        \cr\cmidrule(r){1-3}\cmidrule(r){4-5}
        \cmidrule(r){6-7}\cmidrule(r){8-9}
        \cmidrule(r){10-11}\cmidrule(r){12-13}
        \cmidrule(r){14-15}\cmidrule(r){16-17}
        $L^s_{id}$&MoAE&HDW&
        R1&mAP&R1&mAP&R1&mAP&R1&mAP&R1&mAP&R1&mAP&R1&mAP
        \cr\midrule
         & & &
        50.90&34.75&95.02&80.23&32.99&21.48&
        74.53&43.84&38.89&23.88&38.05&23.74&
        25.92&15.31\cr
        \Checkmark& & &
        52.03&37.49&96.95&85.61&33.68&22.17&
        78.71&50.23&38.64&24.61&40.78&27.62&
        23.44&14.71\cr
        \Checkmark&\Checkmark& &
        53.70&38.76&97.04&86.09&35.31&23.19&
        79.35&50.89&38.19&24.86&45.16&30.95&
        27.15&16.58\cr
        \Checkmark& &\Checkmark&
        53.32&39.61&97.44&86.77&33.14&22.86&
        79.72&52.44&36.89&25.76&46.61&32.50&
        26.15&17.32\cr
        \Checkmark&\Checkmark&\Checkmark&
        \textbf{55.80}&\textbf{41.38}&
        \textbf{97.76}&\textbf{87.97}&
        \textbf{36.75}&\textbf{25.89}&
        \textbf{81.32}&\textbf{53.82}&
        \textbf{39.54}&\textbf{26.93}&
        \textbf{50.25}&\textbf{34.94}&
        \textbf{29.21}&\textbf{18.71}\cr
        \bottomrule[1pt]
    \end{tabular}
    \vspace{-0.1cm}
    \caption{
    Ablation study on AT-USTC.
    Rank-1 (R1) and mAP accuracy (\%) are reported.
    }
    \vspace{-0.2cm}
    \label{tab:abl}
\end{table*}

\begin{table*}[t]
    \renewcommand{\arraystretch}{1.4}      
    \fontsize{9}{9}\selectfont     
    \setlength\tabcolsep{6.0pt}
    \centering
    \begin{tabular}{ccccccccccccccc}
        \toprule[1pt]
        \multicolumn{3}{c}{Method}&
        \multicolumn{2}{c}{\cellcolor[HTML]{C9FAFC}{Avg}}&
        \multicolumn{2}{c}{\cellcolor[HTML]{F5F5CA}{Market1501}}&
        \multicolumn{2}{c}{\cellcolor[HTML]{F5F5CA}{CUHK03}}&
        \multicolumn{2}{c}{\cellcolor[HTML]{FACCCE}{SYSU-MM01}}&
        \multicolumn{2}{c}{\cellcolor[HTML]{FAE2CC}{PRCC}}&
        \multicolumn{2}{c}{\cellcolor[HTML]{FAE2CC}{LTCC}}
        \cr\cmidrule(r){1-3}\cmidrule(r){4-5}
        \cmidrule(r){6-7}\cmidrule(r){8-9}
        \cmidrule(r){10-11}\cmidrule(r){12-13}
        \cmidrule(r){14-15}
        $L^s_{id}$&MoAE&HDW&
        R1&mAP&R1&mAP&R1&mAP&R1&mAP&R1&mAP&R1&mAP
        \cr\midrule
         & & &
        31.47&24.16&60.01&34.30&20.43&19.30&
        20.78&20.81&33.95&33.63&22.19&12.75 
        \cr
        \Checkmark& & &
        34.42&27.21&63.24&37.76&24.79&23.79&
        26.10&25.68&32.97&35.09&25.00&13.72
        \cr
        \Checkmark&\Checkmark& &
        36.95&29.11&68.38&42.87&25.57&24.67&
        28.58&28.03&35.70&35.41&26.53&14.58
        \cr
        \Checkmark& &\Checkmark&
        38.13&30.03&68.82&43.08&27.21&25.99&
        30.21&29.28&35.85&36.88&\textbf{28.57}&14.90
        \cr
        \Checkmark&\Checkmark&\Checkmark&
        \textbf{39.81}&\textbf{31.76}&
        \textbf{70.72}&\textbf{45.67}&
        \textbf{29.07}&\textbf{27.49}&
        \textbf{32.26}&\textbf{31.12}&
        \textbf{38.67}&\textbf{38.88}&
        28.32&\textbf{15.66}
        \cr
        \bottomrule[1pt]
    \end{tabular}
    \vspace{-0.1cm}
    \caption{
    Ablation study with cross-dataset testing. Trained on AT-USTC and inferred on Market1501, CUHK03, SYSU-MM01, PRCC and LTCC datasets.
    Rank-1 (R1) and mAP accuracy (\%) are reported.
    }
    \vspace{-0.2cm}
    \label{tab:abl2}
\end{table*}

\begin{table*}[t]
    \renewcommand{\arraystretch}{1.4}           
    \fontsize{9}{9}\selectfont          
    \setlength\tabcolsep{3.5pt}
    \centering
    \begin{tabular}{l l cc cc cc cc cc cc cc}
        \toprule[1pt]
        &&
        \multicolumn{2}{c}{\cellcolor[HTML]{C9FAFC}{Any-Time}}&
        \multicolumn{2}{c}{\cellcolor[HTML]{F5F5CA}{DT-ST}}&
        \multicolumn{2}{c}{\cellcolor[HTML]{FAE2CC}{DT-LT}}&
        \multicolumn{2}{c}{\cellcolor[HTML]{C5B7D8}{NT-ST}}&
        \multicolumn{2}{c}{\cellcolor[HTML]{9EBFE3}{NT-LT}}&
        \multicolumn{2}{c}{\cellcolor[HTML]{FACCCE}{AD-ST}}&
        \multicolumn{2}{c}{\cellcolor[HTML]{E9BCD9}{AD-LT}}
        \cr\cmidrule(r){3-4}\cmidrule(r){5-6}
        \cmidrule(r){7-8}\cmidrule(r){9-10}
        \cmidrule(r){11-12}\cmidrule(r){13-14}
        \cmidrule(r){15-16}
        \multirow{-2}*{Task}&
        \multirow{-2}*{Method}
        &R1&mAP&R1&mAP&R1&mAP&R1&mAP&R1&mAP&R1&mAP&R1&mAP
        \cr\midrule
        \cellcolor[HTML]{F5F5CA}{}
        &BoT~\cite{luo2019bag}&
        44.94&22.56&89.32&57.03&32.15&14.93&
        63.69&26.72&33.75&15.25&29.78&12.87&
        20.93&8.55
        \cr       
        \cellcolor[HTML]{F5F5CA}{}
        &TransReID~\cite{he2021transreid}&
        48.30&31.42&93.55&75.35&34.17&21.83&
        68.79&36.98&36.50&22.94&32.45&17.72&
        24.34&13.69
        \cr
        \cellcolor[HTML]{F5F5CA}{}
        &CLIP-ReID (R50)~\cite{li2023clip}&
        48.29&29.37&92.00&66.50&30.17&19.08&
        69.71&36.50&36.79&19.54&37.61&20.62&
        23.46&13.96
        \cr
        \multirow{-4}{*}{\cellcolor[HTML]{F5F5CA}{\rotatebox{90}{\shortstack{Tr-\\ReID}}}}
        &CLIP-ReID (ViT-B)&
        52.56&35.86&96.35&81.21&\textbf{41.15}&\textbf{27.34}&
        72.14&41.49&36.00&21.69&42.28&26.30&
        27.43&17.14   
        \cr\midrule
        \cellcolor[HTML]{FAE2CC}{}
        &CAL~\cite{gu2022clothes}&
        50.53&33.95&94.53&76.92&31.80&20.19&
        73.15&41.98&30.60&19.30&47.18&29.75&
        25.92&15.55 
        \cr
        \cellcolor[HTML]{FAE2CC}{}
        &AIM~\cite{yang2023good}&
        50.19&33.31&94.16&76.00&30.37&19.30&
        72.87&41.35&31.90&19.68&46.34&28.69&
        25.52&14.88
        \cr
        \multirow{-3}{*}{\cellcolor[HTML]{FAE2CC}{\rotatebox{90}{\shortstack{CC-\\ReID}}}}
        &CCIL~\cite{li2023clothes}&
        51.58&31.74&92.89&72.11&36.65&23.29&
        70.23&34.89&38.89&22.72&41.43&21.47&
        29.41&15.96
        \cr\midrule

        \cellcolor[HTML]{FACCCE}{}
        &CAJ~\cite{ye2021channel}&
        50.09&30.04&93.38&68.91&34.17&20.83&
        66.76&33.15&36.10&20.07&41.79&22.21&
        28.33&15.05
        \cr
        \cellcolor[HTML]{FACCCE}{}
        &CIFT$^\dagger$~\cite{li2022counterfactual}&
        53.29&33.47&92.32&72.88&38.92&24.56&
        71.77&36.92&38.04&22.68&48.61&26.20&
        30.11&17.61
        \cr
        \multirow{-3}{*}{\cellcolor[HTML]{FACCCE}{\rotatebox{90}{\shortstack{CM-\\ReID}}}}
        &DEEN~\cite{zhang2023diverse}&
        53.52&33.48&91.86&70.11&37.34&24.47&
        71.34&37.76&38.54&22.83&50.06&27.91&
        \textbf{31.66}&18.09
        \cr\midrule

        \cellcolor[HTML]{C9FAFC}{}
        &Baseline&
        50.90&34.75&95.02&80.23&32.99&21.48&
        74.53&43.84&38.89&23.88&38.05&23.74&
        25.92&15.31
        \cr
        \multirow{-2}{*}{\cellcolor[HTML]{C9FAFC}{\rotatebox{90}{\shortstack{AT-\\ReID}}}}
        &Uni-AT (Ours)&
        \textbf{55.80}&\textbf{41.38}&
        \textbf{97.76}&\textbf{87.97}&
        36.75&25.89&
        \textbf{81.32}&\textbf{53.82}&
        \textbf{39.54}&\textbf{26.93}&
        \textbf{50.25}&\textbf{34.94}&
        29.21&\textbf{18.71}\cr
        \bottomrule[1pt]
    \end{tabular}
    \vspace{-0.1cm}
    \caption{
    Comparison with task-specific methods on AT-USTC.
    Rank-1 (R1) and mAP accuracy (\%) are reported.
    }
    \vspace{-0.1cm}
    \label{tab:comp}
\end{table*}

\begin{table*}[t]
    \renewcommand{\arraystretch}{1.4}
    \fontsize{9}{9}\selectfont
    \setlength\tabcolsep{5.2pt}
    \centering
    \begin{tabular}{l l cc cc cc cc cc cc}
        \toprule[1pt]
        &&
        \multicolumn{2}{c}{\cellcolor[HTML]{C9FAFC}{Avg}}&
        \multicolumn{2}{c}{\cellcolor[HTML]{F5F5CA}{Market1501}}&
        \multicolumn{2}{c}{\cellcolor[HTML]{F5F5CA}{CUHK03}}&
        \multicolumn{2}{c}{\cellcolor[HTML]{FACCCE}{SYSU-MM01}}&
        \multicolumn{2}{c}{\cellcolor[HTML]{FAE2CC}{PRCC}}&
        \multicolumn{2}{c}{\cellcolor[HTML]{FAE2CC}{LTCC}}
        \cr\cmidrule(r){3-4}\cmidrule(r){5-6}
        \cmidrule(r){7-8}\cmidrule(r){9-10}
        \cmidrule(r){11-12}\cmidrule(r){13-14}
        \multirow{-2}*{Task}&
        \multirow{-2}*{Method}
        &R1&mAP&R1&mAP&R1&mAP&R1&mAP&R1&mAP&R1&mAP
        \cr\midrule
        \cellcolor[HTML]{F5F5CA}{}
        &BoT~\cite{luo2019bag}&
        27.89&16.00&50.06&21.40&10.93&9.25&
        15.42&14.47&36.78&26.78&26.28&8.11
        \cr      
        \cellcolor[HTML]{F5F5CA}{}
        &TransReID~\cite{he2021transreid}&
        30.19&22.02&57.75&31.04&19.36&18.00&
        14.36&14.85&34.97&34.96&24.49&11.25
        \cr
        \cellcolor[HTML]{F5F5CA}{}
        &CLIP-ReID (R50)~\cite{li2023clip}&
        28.22&18.36&46.32&20.32&11.50&10.56&
        19.51&19.19&35.73&30.50&28.06&11.22
        \cr
        \cellcolor[HTML]{F5F5CA}{}
        \multirow{-4}{*}{\cellcolor[HTML]{F5F5CA}{\shortstack{Tr-\\ReID}}}
        &CLIP-ReID (ViT-B)&
        \textbf{42.97}&31.68&\textbf{73.90}&\textbf{47.73}&27.93&24.52&
        29.28&28.07&\textbf{45.30}&\textbf{40.83}&\textbf{38.52}&\textbf{17.24}
        \cr\midrule

        \cellcolor[HTML]{FAE2CC}{}
        &CAL~\cite{gu2022clothes}&
        34.33&24.74&59.06&32.56&19.71&19.34&
        27.09&26.16&36.21&33.63&29.59&12.03
        \cr
        \cellcolor[HTML]{FAE2CC}{}
        &AIM~\cite{yang2023good}&
        33.94&24.66&59.92&32.76&20.07&19.37&
        25.76&25.22&35.39&33.67&28.57&12.30
        \cr
        \multirow{-3}{*}{\cellcolor[HTML]{FAE2CC}{\shortstack{CC-\\ReID}}}
        &CCIL~\cite{li2023clothes}&
        31.30&21.92&54.72&27.39&14.50&13.79&
        20.09&18.82&42.70&38.74&24.49&10.87
        \cr\midrule

        \cellcolor[HTML]{FACCCE}{}
        &CAJ~\cite{ye2021channel}&
        31.33&21.73&55.11&27.77&16.36&15.14&
        23.58&22.15&35.06&32.82&26.53&10.75
        \cr
        \cellcolor[HTML]{FACCCE}{}
        &CIFT$^\dagger$~\cite{li2022counterfactual}&
        32.14&22.78&54.78&27.57&13.79&14.18&
        24.00&22.66&41.86&38.79&26.28&10.70
        \cr
        \multirow{-3}{*}{\cellcolor[HTML]{FACCCE}{\shortstack{CM-\\ReID}}}
        &DEEN~\cite{zhang2023diverse}&
        31.66&21.49&55.31&27.47&14.93&13.58&
        26.78&24.73&35.79&31.42&25.51&10.27
        \cr\midrule

        \cellcolor[HTML]{C9FAFC}{}
        &Baseline&
        31.47&24.16&60.01&34.30&20.43&19.30&
        20.78&20.81&33.95&33.63&22.19&12.75 
        \cr
        \multirow{-2}{*}{\cellcolor[HTML]{C9FAFC}{\shortstack{AT-\\ReID}}}
        &Uni-AT (Ours)&
        39.81&\textbf{31.76}&70.72&45.67&\textbf{29.07}&\textbf{27.49}&         \textbf{32.26}&\textbf{31.12}&38.67&38.88&28.32&15.66
        \cr
        \bottomrule[1pt]
    \end{tabular}
    \vspace{-0.1cm}
    \caption{
    Comparison with task-specific methods with cross-dataset testing. Trained on AT-USTC and inferred on Market1501, CUHK03, SYSU-MM01, PRCC and LTCC datasets.
    Rank-1 (R1) and mAP accuracy (\%) are reported.
    }
    \vspace{-0.1cm}
    \label{tab:comp2}
\end{table*}

\paragraph{Implementation Details.}
We use a ViT-Base model with patch size 16 and step size 16 as our backbone.
Following existing work~\cite{luo2019bag}, we introduce the BNNeck before the classifier.
All person images are resized to 256 × 128 and are augmented with random horizontal flipping, padding, random cropping, and random erasing~\cite{zhong2020random} in training.
The batch size is set to 64 with 8 identities.
The whole model is trained for 120 epochs (24K iterations) with the SGD optimizer. 
The learning rate is initialized as 0.008 with the warm-up scheme and cosine learning rate decay.

For the other comparison methods, we employed their official code, ensuring that the image are resized to 256 × 128. For the transformer-based methods, we maintained the use of the ViT-Base model with a patch size of 16 and a step size of 16.

\subsection{Generalization Evaluation of AT-USTC}
One of our main contributions is collecting the AT-USTC dataset, which exhibits higher intra-identity diversity compared to existing ReID datasets.
To evaluate the quality of our dataset, we conducted domain generalization experiments, comparing it with three large-scale Tr-ReID, CC-ReID, and CM-ReID datasets, including MSMT17, DeepChange, and LLCM.
For a fair comparison at the dataset level, we trained the same ResNet50 model for all datasets instead of our proposed model. 

As shown in Tab.~\ref{tab:dataset}, the model trained on AT-USTC achieved the best cross-dataset performance, surpassing the model trained on MSMT17/DeepChange/LLCM by an average of 10.90\% / 13.69\% / 14.36\% in Rank-1 accuracy and 9.23\% / 11.26\% / 12.64\% in mAP accuracy.
This indicates that, in addition to the number of IDs, the inherent diversity of each ID is a significant aspect of the ReID dataset. The high intra-identity diversity in AT-USTC results in excellent scalability, thereby effectively supporting research on the AT ReID task

\subsection{Comparison with MoE Methods}
As shown in Tab.~\ref{tab:moe}, we compare our MoAE module with other MoE methods under Any-Time testing on the AT-USTC dataset.
For a fair comparison, all MoE modules are added to our MS-ReID framework in the same manner.
The difference among these methods lies in the sharing of experts.
MMoE~\cite{ma2018modeling} constructs experts shared across all scenarios, while PLE~\cite{tang2020progressive} explicitly separates the experts into scenario-specific ones and those shared across all scenarios. 
VLMo~\cite{bao2022vlmo} was originally designed to process visual and language modalities. 
Here, we apply its principles to handle the RGB and IR modalities in the ReID task, establishing three types of experts for the RGB, IR, and cross-modality data.

From the experimental results, it can be summarized that our MoAE consistently outperforms other methods when parameters or computational time are comparable.
This is because our MoAE introduces scenario priors and enables fine-grained expert sharing, whereas other methods can only coarsely share experts across scenarios.
Furthermore, our training or inference speed is only 1.2 times that of the non-expert model when using 12 experts.
Through parameter sharing with our attribute layers, MoAE not only effectively extracts scenario-specific features to enhance performance but also exhibits greater parameter and time efficiency than other MoE methods.

\subsection{Ablation Analysis}
As shown in Tab.~\ref{tab:abl} and Tab.~\ref{tab:abl2}, we conduct ablation studies to evaluate each component of our method.
In the 1-$st$ row, we establish an MS-ReID baseline using the standard identity loss for six CLS tokens, which provides undifferentiated supervision for all scenarios, failing to take advantage of our MS-ReID framework. 
In the 2-$nd$ row, we use scenario-aware identity loss $L^s_{id}$ to facilitate the effective knowledge acquisition from each scenario, leading to improvements of 1.13\% and 2.74\% in Rank-1 and mAP accuracy on Any-Time testing.
However, this approach tackles each scenario independently without considering their relationships, limiting efficient collaborative optimization across multiple scenarios.

To address this issue, we further introduce the MoAE and HDW components displayed in the 3-$rd$ and 4-$th$ rows.
The integration of MoAE results in notable improvements of 1.67\% and 1.77\% in Rank-1 and mAP accuracy.
The MoAE module introduces an expert mechanism to guide the model in capturing discriminative scenario-specific cues, effectively improving feature learning across relevant scenarios.
Furthermore, the incorporation of the HDW scheme improves Rank-1 and mAP accuracy by 1.29\% and 2.12\%.
This scheme facilitates balanced learning of specific features across various scenarios in an end-to-end manner.
Ultimately, as shown in the 5-$th$ row, the combination of all components leads to remarkable performance enhancements of 4.90\% and 6.63\% in Rank-1 and mAP accuracy on Any-Time testing, and 10.71\% / 8.64\% / 11.58\% and 4.72\% / 6.13\% improvements in Rank-1 accuracy for cross-dataset testing on the Market1501 / CUHK03 / SYSU-MM01 / PRCC / LTCC dataset.
This demonstrates the complementary of the proposed $L^s_{id}$, MoAE, and HDW, making the entire MS-ReID framework more effective.

\subsection{Comparison with ReID Methods}

As shown in Tab.~\ref{tab:comp}, we compare our method with several popular task-specific methods of Tr-ReID, CC-ReID, and CM-ReID tasks.
Our method achieves the highest performance, surpassing CLIP-ReID~\cite{li2023clip} by 3.14\% Rank-1 and 5.52\% mAP accuracy on Any-Time testing.
Our method also surpasses the methods of CC-ReID and CM-ReID on the six AT-ReID scenarios.
This demonstrates that existing methods designed for a target scenario are inadequate for addressing the challenges of the multi-scenario AT-ReID.
These methods learn a unified representation for all scenarios and assign all images of a person to a single category regardless of modality or clothing. 
Consequently, they focus solely on shared features across six scenarios while neglecting specific cues, resulting in poor generalization in non-target scenarios and diminished retrieval performance in target scenarios.

Furthermore, as shown in Tab.~\ref{tab:comp2}, our AT-USTC dataset and Uni-AT method demonstrate strong generalization capabilities, exhibiting excellent performance in cross-dataset testing.
Our Uni-AT approach employs a novel multi-scenario learning framework to capture specific features of different scenarios and facilitate mutual enhancement among scenarios, which exhibits notable advantages compared to the baseline and other ReID methods.

\section{Additional Analysis}
\label{sec:additional_analysis}

In this section, we provide additional analysis for the proposed Uni-AT model.
The goal is to better understand the training dynamics of multi-scenario learning, including the behavior of HDW, the expert routing of MoAE, and the inter-scenario gradient conflict.

\subsection{Analysis of Hierarchical Dynamic Weighting}
\label{sec:analysis_hdw}

\begin{figure*}[!t]
    \centering
    \begin{minipage}[t]{0.49\linewidth}
        \centering
        \includegraphics[width=\linewidth]{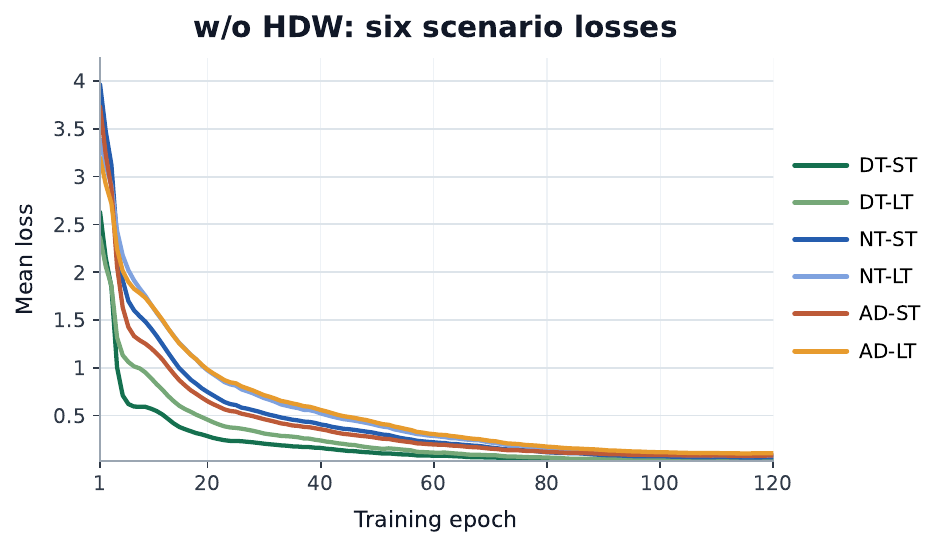}
        {\small (a) Scenario losses without HDW.}
    \end{minipage}
    \hfill
    \begin{minipage}[t]{0.49\linewidth}
        \centering
        \includegraphics[width=\linewidth]{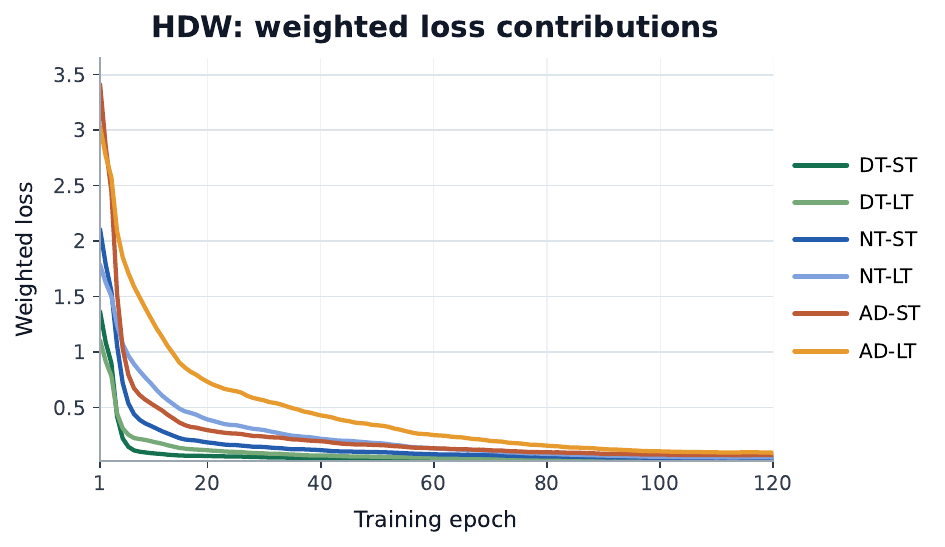}
        {\small (b) Weighted loss contributions with HDW.}
    \end{minipage}

    \vspace{0.15cm}
    \begin{minipage}[t]{0.49\linewidth}
        \centering
        \includegraphics[width=\linewidth]{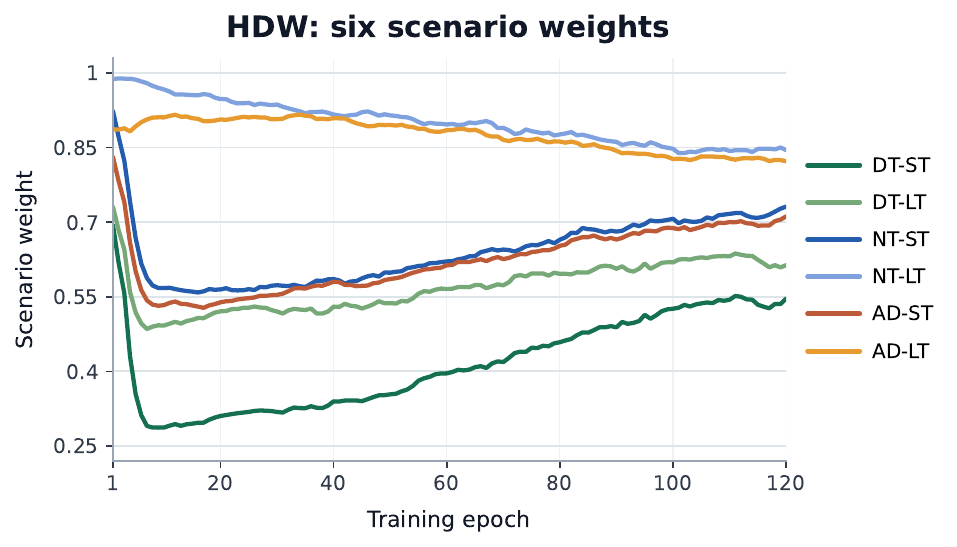}
        {\small (c) Scenario-level HDW weights.}
    \end{minipage}
    \hfill
    \begin{minipage}[t]{0.49\linewidth}
        \centering
        \includegraphics[width=\linewidth]{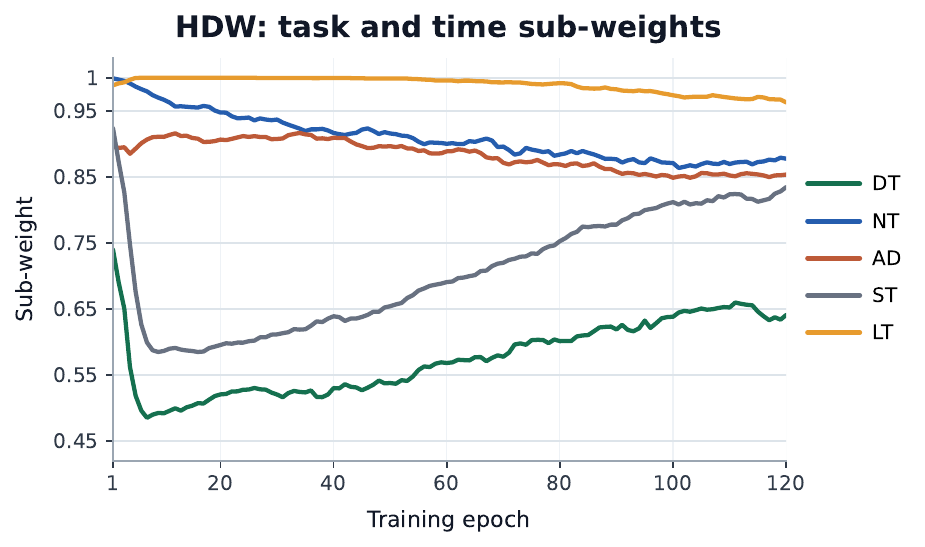}
        {\small (d) Task and time sub-weights.}
    \end{minipage}
    \vspace{-0.2cm}
    \caption{Analysis of HDW training dynamics on AT-USTC. We visualize the original scenario losses, the weighted loss contributions, the scenario-level weights, and the task/time sub-weights during training.}
    \vspace{-0.2cm}
    \label{fig:analysis_hdw}
\end{figure*}

Fig.~\ref{fig:analysis_hdw} visualizes how HDW adjusts the optimization process of six AT-ReID scenarios.
Without HDW, the six scenario losses have different magnitudes and learning curves.
For example, the all-day and long-term related scenarios tend to keep larger losses for a longer period, while some short-term scenarios converge faster.
This indicates that simply summing all scenario losses may let easier tasks dominate part of the optimization and may provide insufficient attention to harder scenarios.

With HDW, the weighted loss contributions become more balanced across scenarios.
This behavior is consistent with the motivation of HDW: the model should dynamically emphasize the scenarios that remain difficult during training while reducing the relative contribution of easier ones.
The scenario-level weights in Fig.~\ref{fig:analysis_hdw}(c) show that HDW does not assign static preferences.
Instead, the weights change with the learning status of different scenarios.
The task and time sub-weights in Fig.~\ref{fig:analysis_hdw}(d) further show that the hierarchical design captures two types of scenario relationships, namely the time-moment relation among DT, NT, and AD, and the time-interval relation between ST and LT.
Therefore, HDW balances the six losses not only at the scenario level, but also through the shared attributes behind these scenarios.

\subsection{Analysis of MoAE Expert Routing}
\label{sec:analysis_moae_routing}

\begin{figure*}[!t]
    \centering
    \begin{minipage}[t]{0.49\linewidth}
        \centering
        \includegraphics[width=\linewidth]{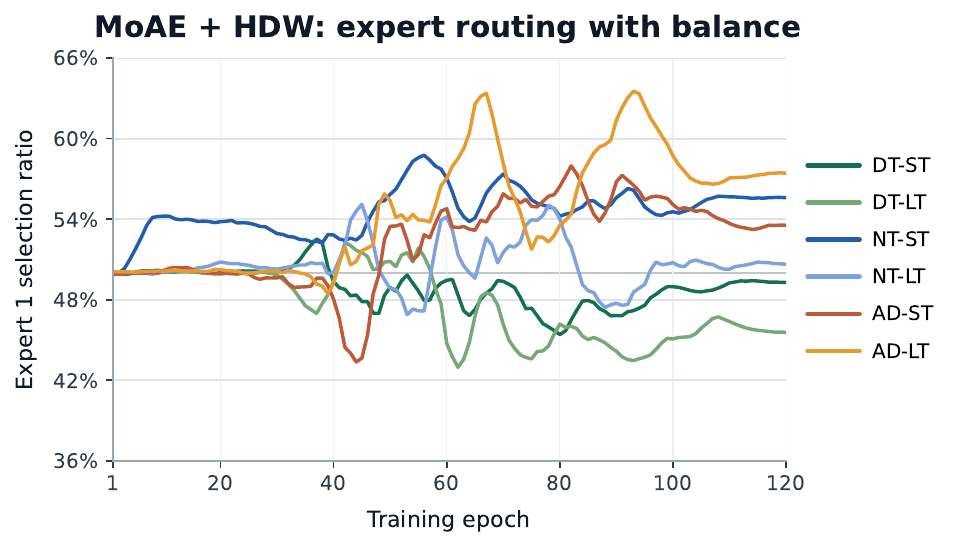}
        {\small (a) With bias update.}
    \end{minipage}
    \hfill
    \begin{minipage}[t]{0.49\linewidth}
        \centering
        \includegraphics[width=\linewidth]{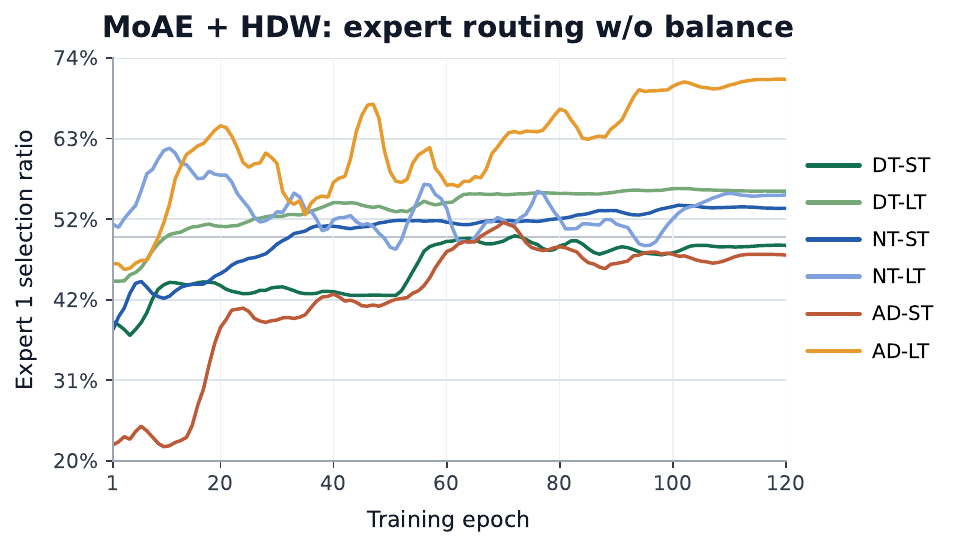}
        {\small (b) Without bias update.}
    \end{minipage}
    \vspace{-0.2cm}
    \caption{Analysis of MoAE expert routing. We compare the selection ratio of one expert branch with and without a routing-bias update.}
    \vspace{-0.2cm}
    \label{fig:analysis_moae_routing}
\end{figure*}

We further analyze whether the experts in MoAE are effectively used.
According to Sec.~\ref{sec:moae}, the original MoAE combines scenario-specific experts by
\begin{equation}
\begin{aligned}
y &= \sum\nolimits_{j=1}^{n} G^s(t_i^s)_j E_j^s(t_i^s),\\
G^s(t_i^s) &= top_k(\mathrm{softmax}(W_g^s\cdot t_i^s)).
\end{aligned}
\end{equation}
To study the effect of explicitly balancing the routing process, we introduce a bias vector $b^s$ into the routing logits:
\begin{equation}
\hat{G}^s(t_i^s) =
top_k(\mathrm{softmax}(W_g^s\cdot t_i^s+b^s)).
\end{equation}
The bias is directly updated according to the observed expert selection ratio.
For each scenario $s$, the selection ratio of the $j$-th expert in a batch $\mathcal{B}_s$ is computed as:
\begin{equation}
r_j^s =
\frac{1}{|\mathcal{B}_s|}
\sum_{t_i^s\in \mathcal{B}_s}
\mathbf{1}\left[\arg\max_m \hat{G}^s(t_i^s)_m=j\right].
\end{equation}
Then the routing bias is updated by
\begin{equation}
b_j^s \leftarrow b_j^s + \eta\left(\frac{1}{n}-r_j^s\right),
\end{equation}
where $\eta$ is the update step size.
This update decreases the routing tendency of over-selected experts and increases that of under-selected experts.

Fig.~\ref{fig:analysis_moae_routing} shows that the bias update can make the expert selection ratios closer to a uniform distribution.
However, MoAE without this explicit update does not collapse to a single expert branch; its routing ratios still remain in a reasonable range for most scenarios.
This behavior is consistent with the design of MoAE.
Although each scenario has its own gate, the experts are constructed from shared time-moment and time-interval attribute layers.
Such attribute sharing naturally regularizes expert usage across related scenarios and prevents severe routing imbalance.
Therefore, MoAE's sharing mechanism is already sufficiently balanced in practice, suggesting that the model does not strongly rely on an additional bias update to maintain reasonable expert utilization.

\subsection{Analysis of Inter-Scenario Gradient Conflict}
\label{sec:analysis_gradient_conflict}

\begin{figure*}[!t]
    \centering
    \includegraphics[width=0.70\linewidth]{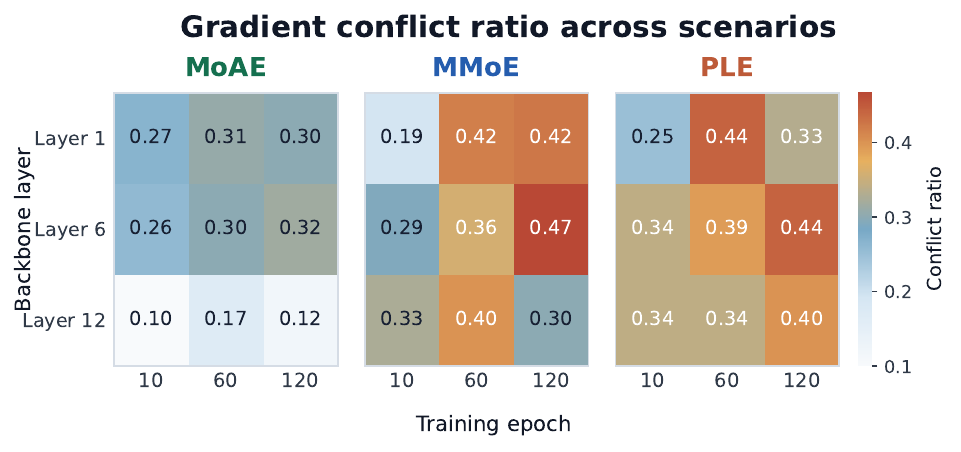}
    \vspace{-0.2cm}
    \caption{Gradient conflict ratio across scenarios. We compare MoAE with representative MoE variants at different backbone layers and training epochs.}
    \vspace{-0.2cm}
    \label{fig:analysis_gradient_conflict}
\end{figure*}

Multi-scenario training naturally introduces different optimization objectives.
For example, DT-related scenarios may rely on RGB-specific cues, NT-related scenarios require infrared-compatible cues, ST-related scenarios can exploit clothing details, and LT-related scenarios should be less sensitive to clothing changes.
These different objectives may cause gradient conflict among scenario losses.
To measure this effect, we compute the gradient conflict ratio across scenario pairs.
For two scenario losses, a conflict is counted when the cosine similarity between their gradients is negative:
\begin{equation}
\mathrm{CR}
=
\frac{1}{|\mathcal{P}|}
\sum_{(a,b)\in\mathcal{P}}
\mathbf{1}\left[\cos(g_a,g_b)<0\right],
\end{equation}
where $g_a$ and $g_b$ are the gradients of two scenario losses with respect to the same backbone layer, and $\mathcal{P}$ is the set of scenario pairs.

As shown in Fig.~\ref{fig:analysis_gradient_conflict}, MoAE generally yields lower or more stable conflict ratios than the compared MoE variants.
The advantage is especially clear in deeper layers, where scenario-specific semantics are more strongly formed.
This result indicates that attribute-aware expert sharing can better organize the learning of different scenarios.
Compared with coarse expert sharing, MoAE provides a more suitable structure for AT-ReID because it separates incompatible scenario-specific cues while still sharing related attribute knowledge.
The gradient analysis is also consistent with the quantitative comparison in Tab.~\ref{tab:moe}, where MoAE achieves better accuracy with lower parameter and time overhead than other MoE designs.

\begin{figure*}[!t]
    \centering
    \includegraphics[width=1.0\linewidth]{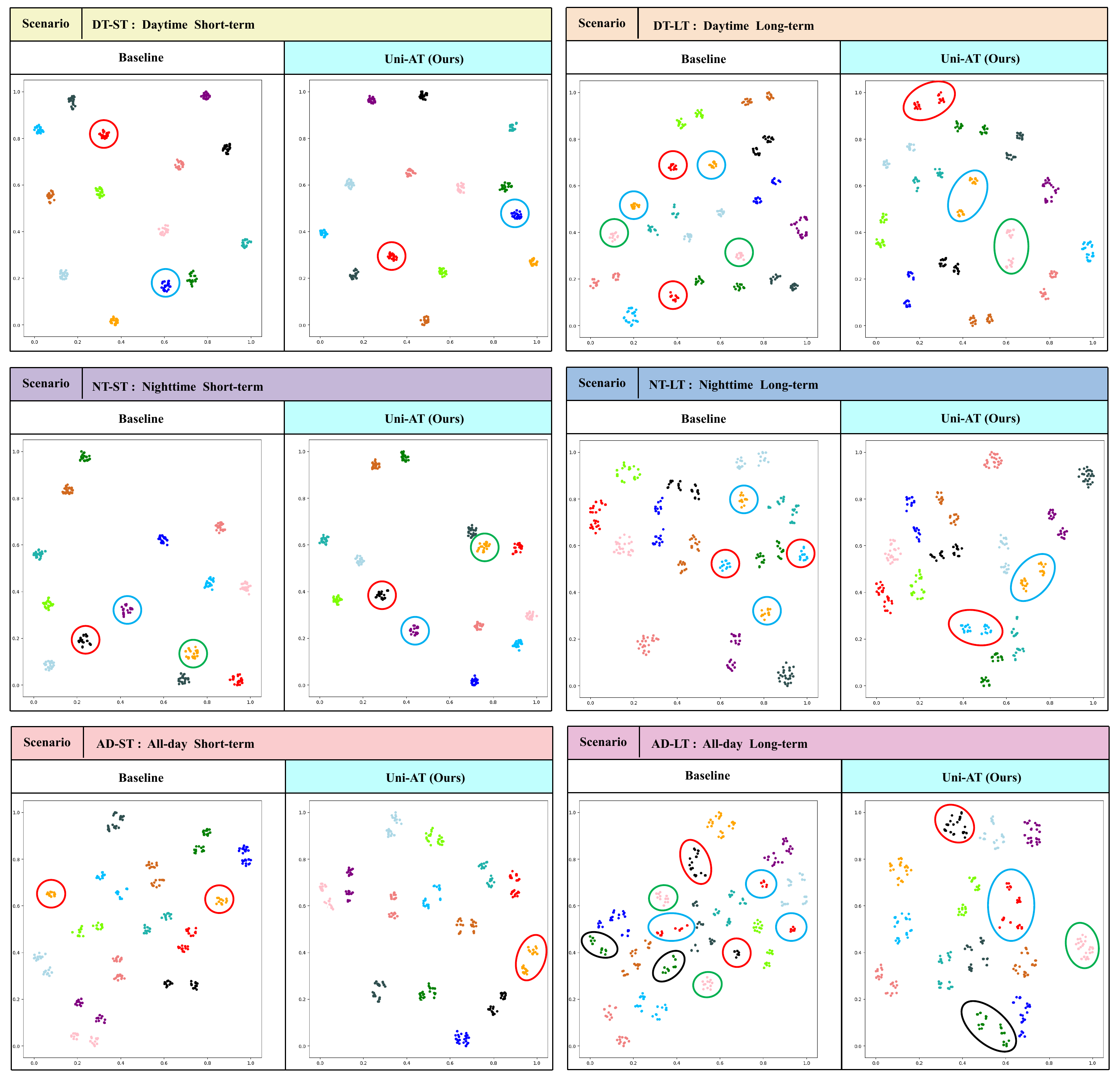}
    \caption{t-SNE visualization of the feature distributions in six AT-ReID scenarios on the AT-USTC dataset. Dots of the same color represent the features of the same person in different images. Circles highlight examples where our approach is better than the baseline.}
    \hfill
    \label{fig:tsne}
\end{figure*}

\begin{figure*}[!t]
    \centering
    \includegraphics[width=1.0\linewidth]{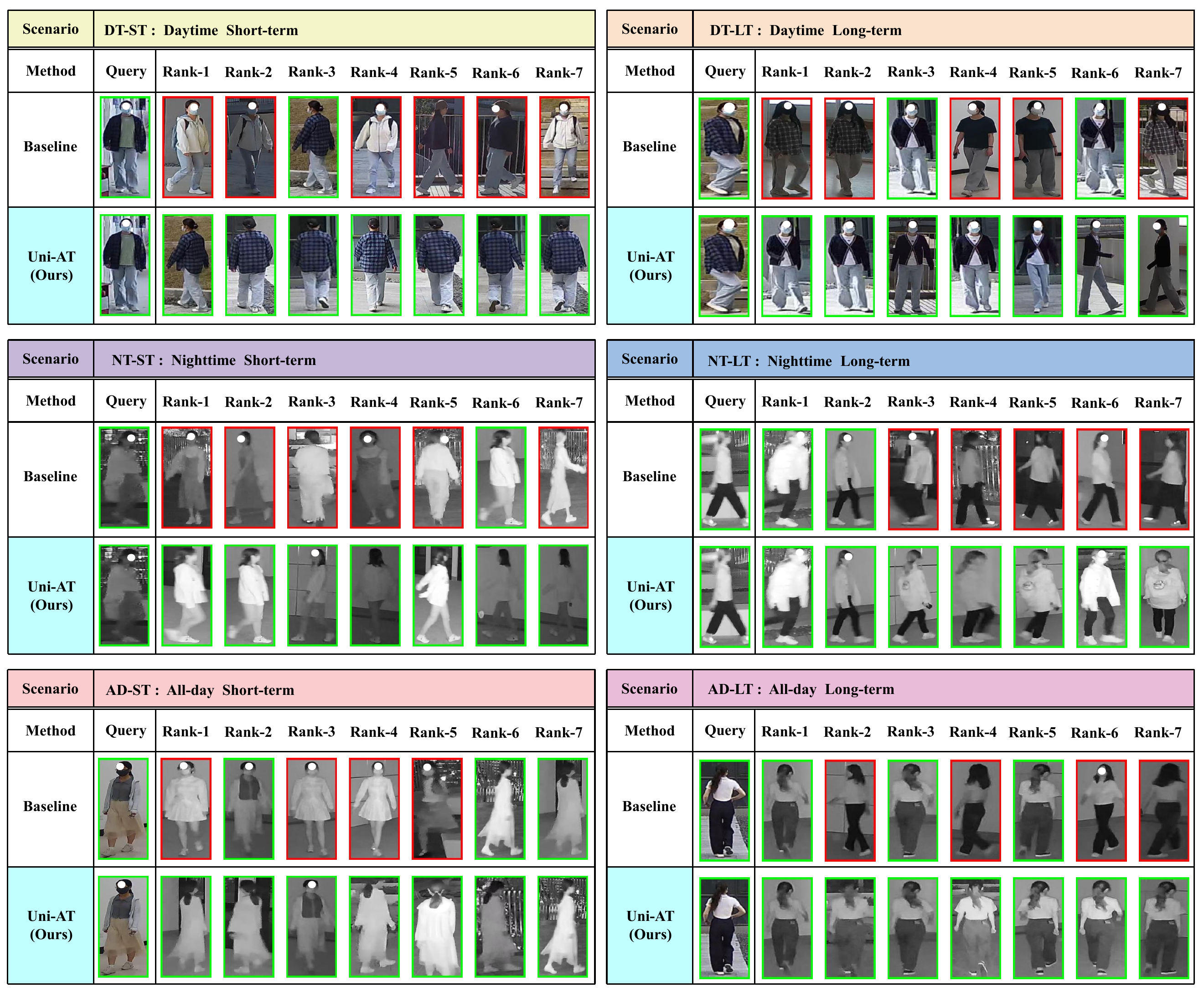}
    \caption{Visualization of the retrieval ranking lists in six AT-ReID scenarios on the AT-USTC dataset. The green boxes represent correct retrieval results, and the red boxes represent incorrect retrieval results. 
    }
    \hfill
    \label{fig:ranking}
\end{figure*}

\section{Visualizations}
In this section, we compared the feature distributions and ranking lists between the baseline model and the proposed Uni-AT model on the subset of the AT-USTC dataset.

\paragraph{Visualization of the feature distribution.}
Fig.~\ref{fig:tsne} shows t-SNE~\cite{van2008visualizing} visualization results in six AT-ReID scenarios. 
Circles highlight examples where our approach is better than the baseline.
In the DT-ST and NT-ST scenarios, our approach has a more compact distribution of intra-class features.
In other complex scenarios with multiple modality images or multiple sets of clothes for each person, the features of different identities are mixed in the baseline method,
while our method still maintains distinguishable features among different identities.
The results demonstrate that our method is capable of learning discriminative features in various scenarios.

\paragraph{Visualization of the retrieval results.}
As shown in Fig.~\ref{fig:ranking}, we present the retrieval ranking lists in six AT-ReID scenarios on the AT-USTC dataset. 
The green boxes represent the positive search results, and the red boxes represent the negative search results.
The retrieval results indicate that our method can retrieve some hard positive targets that the baseline method cannot.
For example, in ST-related scenarios, our approach leverages clothing information to identify individuals with similar appearances.
Moreover, the availability of clothing information, including style, color, and texture, varies across distinct short-term scenarios.
In contrast, in LT-related scenarios, our method avoids relying on clothing details, which can prevent misidentifying individuals with similar clothing.
The baseline method aims to learn shared representations across all scenarios, thus rendering it incapable of utilizing the specific information of each scenario and leading to avoidable errors.
The results indicate that our Uni-AT has better retrieval capability in various scenarios than the baseline method.

\section{Conclusion}
In this paper, we investigate a new ReID task, Anytime ReID (AT-ReID), and introduce the first corresponding dataset named AT-USTC to support its research.
Compared to existing datasets, AT-USTC is characterized by its extensive intra-identity diversity in clothing, modality, and camera, effectively fulfilling the crucial requirements of AT-ReID.
We analyze this new challenge task and propose a Unified AT-ReID model to handle the multi-scenario AT-ReID through a single model.
By improving model architecture and optimization, we achieve accurate retrieval results across multiple scenarios. 
The proposed methods and datasets may inspire the following ReID methods towards real application.

\section*{Acknowledgements} 
This work is supported by the National Natural Science Foundation of China (Grant No. 62272430). 
\section*{Contribution Statement} 
Xulin Li and Yan Lu made equal contributions to this work.

{
    \small
    \bibliographystyle{ieeenat_fullname}
    \bibliography{main}
}


\end{document}